\documentclass{article} 
\usepackage[preprint]{colm2026_conference}

\usepackage{microtype}
\usepackage{hyperref}
\usepackage{url}
\usepackage{booktabs}
\usepackage{float}

\usepackage{lineno}
\usepackage{multirow}

\definecolor{darkblue}{rgb}{0, 0, 0.5}
\hypersetup{colorlinks=true, citecolor=darkblue, linkcolor=darkblue, urlcolor=darkblue}

\usepackage[accsupp]{axessibility}
\usepackage{multirow}
\usepackage{graphicx}

\usepackage[dvipsnames]{xcolor}
\usepackage{tcolorbox}
\definecolor{vanillacol}{HTML}{E6F0FF}
\definecolor{cotcol}{HTML}{FFF9DB}

\newcommand{\writinginstr}[1]{\colorbox{yellow!25}{\parbox{\dimexpr\linewidth-2\fboxsep}{#1}}}
\newcommand{\norestrictioninstr}[1]{\colorbox{green!20}{\parbox{\dimexpr\linewidth-2\fboxsep}{#1}}}

\newcommand{\inputinstr}[1]{\colorbox{yellow!25}{\parbox{\dimexpr\linewidth-2\fboxsep}{#1}}}
\newcommand{\groundtruthinstr}[1]{\colorbox{green!20}
{\parbox{\dimexpr\linewidth-2\fboxsep}{#1}}}
\newcommand{\generatedinstr}[1]{\colorbox{blue!20}
{\parbox{\dimexpr\linewidth-2\fboxsep}{#1}}}

\newcommand{\sys}{MedConclusion}

\usepackage[table]{xcolor}
\usepackage{pifont}

\newcommand{\cmark}{\textcolor{green!45!black}{\ding{51}}}
\newcommand{\xmark}{\textcolor{red!75!black}{\ding{55}}}

\usepackage{makecell}
\usepackage{multicol}
\usepackage{subcaption} 

\usepackage{silence}
\newcommand{\circled}[1]{%
  \begin{tikzpicture}[baseline=(char.base)]
    \node[shape=circle, draw, inner sep=0.5pt, fill=black, text=white, font=\bfseries\tiny] (char) {#1};
  \end{tikzpicture}%
}

\title{\sys{}: A Benchmark for Biomedical Conclusion Generation from Structured Abstracts}

\newcommand{\harvard}{%
    \includegraphics[width=0.016\textwidth]{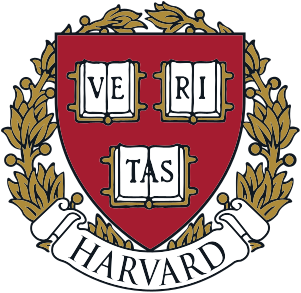} 
}
\newcommand{\stanford}{%
    \includegraphics[width=0.014\textwidth]{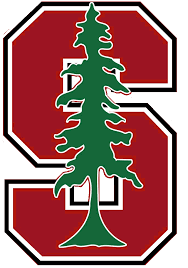} 
}
\newcommand{\cmu}{%
    \includegraphics[width=0.014\textwidth]{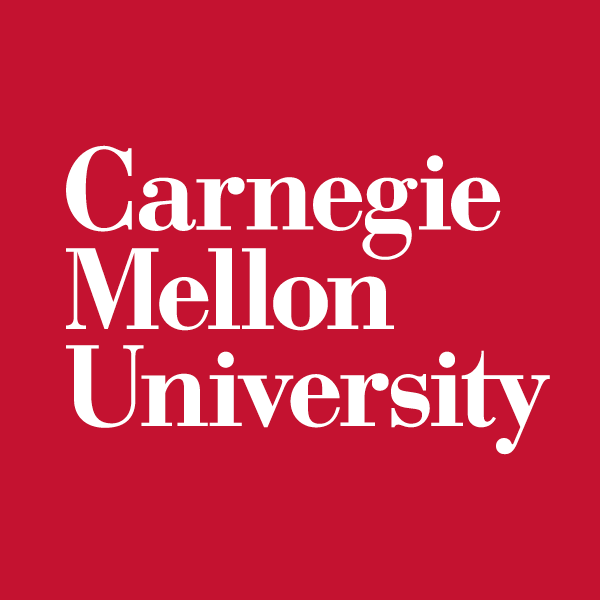} 
}

\newcommand{\usc}{%
    \includegraphics[width=0.018\textwidth]{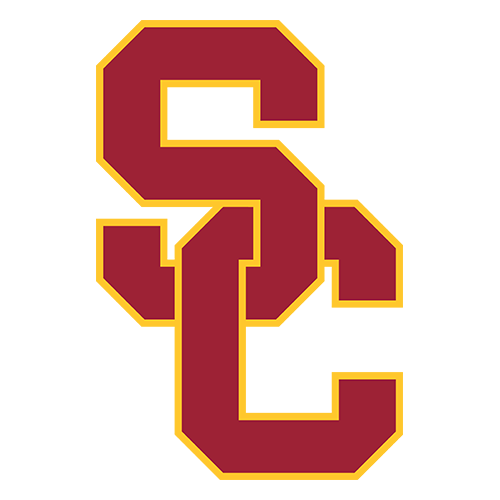} 
}

\newcommand{\kempner}{%
    \includegraphics[width=0.016\textwidth]{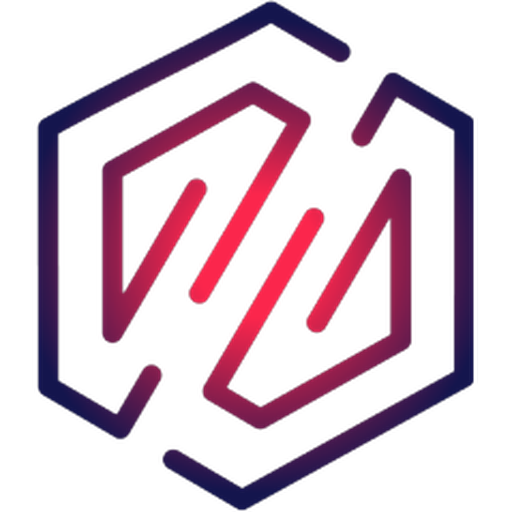} 
}

\newcommand{\squad}{\hspace{0.5em}}
\newcommand{\aff}[1]{\textsuperscript{\normalfont#1}}

\author{\textbf{Weiyue Li}\thanks{Contributed equally as co-first authors.}\squad\aff{\harvard} \squad
\textbf{Ruizhi Qian}\footnotemark[1]\squad\aff{\usc,\harvard} \squad
\textbf{Yi Li}\footnotemark[1]\squad\aff{\cmu,\harvard} \squad
\textbf{Yongce Li}\aff{\stanford} \squad
\textbf{Yunfan Long}\aff{\cmu} \squad
\textbf{Jiahui Cai}\aff{\harvard}\\
\textbf{Yan Luo}\thanks{Contributed equally as co-senior authors.}\squad\aff{\harvard} \squad
\textbf{Mengyu Wang}\footnotemark[2]\squad\aff{\harvard,\kempner} \\
\harvard Harvard AI and Robotics Lab, Harvard Medical School\\
\usc University of Southern California,
\cmu Carnegie Mellon University,
\stanford Stanford University\\
\kempner Kempner Institute for the Study of Natural and Artificial Intelligence, Harvard University
}

\begin{document}

\ifcolmsubmission
\linenumbers
\fi

\maketitle

\begin{abstract}
Large language models (LLMs) are widely explored for reasoning-intensive research tasks, yet resources for testing whether they can infer scientific conclusions from structured biomedical evidence remain limited. We introduce \textbf{\sys{}}, a large-scale dataset of \textbf{5.7M} PubMed structured abstracts for biomedical conclusion generation. Each instance pairs the non-conclusion sections of an abstract with the original author-written conclusion, providing naturally occurring supervision for evidence-to-conclusion reasoning. \sys{} also includes journal-level metadata such as biomedical category and SJR, enabling subgroup analysis across biomedical domains. As an initial study, we evaluate diverse LLMs under conclusion and summary prompting settings and score outputs with both reference-based metrics and LLM-as-a-judge. We find that conclusion writing is behaviorally distinct from summary writing, strong models remain closely clustered under current automatic metrics, and judge identity can substantially shift absolute scores. \sys{} provides a reusable data resource for studying scientific evidence-to-conclusion reasoning. Our code and data are available at: \url{https://github.com/Harvard-AI-and-Robotics-Lab/MedConclusion}.
\end{abstract}

\section{Introduction}

Large language models (LLMs) have shown strong reasoning capability across a wide range of demanding settings, including mathematical thinking, long-form creative writing, and scientific discovery assistance~\citep{luo2025llm4sr,zheng2025automation,li2026llm,zhang2025realmath,yu2025formalmath,fein2026litbench}. As these capabilities improve, there is growing interest in using LLMs to support research workflows, not only to retrieve or summarize papers, but also to infer scientific conclusions from evidence. Structured abstracts provide a particularly convenient setting for studying this capability as we can frame this evidence-to-conclusion reasoning problem as: given \textbf{Background}, \textbf{Methods}, and \textbf{Results}, the model should infer a \textbf{Conclusion} without injecting unprovided context. However, finding large-scale data sources from diverse scientific domains is challenging, which motivates us to shift the focus to biomedicine, where structured abstracts are widely used.

Existing work has explored this direction, but current resources remain limited in two ways. First, existing \emph{datasets} are often narrow in scope, focusing on specific study types or specialized report formats, such as randomized controlled trial abstracts or echocardiography notes, rather than broad biomedical literature \citep{shieh-etal-2019-towards,tang-etal-2022-echogen}. Other work uses conclusion reconstruction mainly as a proxy for premise--conclusion alignment or as a training objective, rather than as a reusable data resource \citep{gao-etal-2024-evaluating,bastan-etal-2022-sume}. These resources also typically do not emphasize journal-level metadata such as biomedical category and SJR, which limits analysis of how difficulty varies across subfields or venue strata. Second, existing \emph{benchmarking designs} do not fully isolate the reasoning problem of conclusion generation. Some adjacent biomedical resources focus on question answering, medical exam reasoning, treatment-effect inference, or claim verification rather than deriving the author-written conclusion itself \citep{jin-etal-2019-pubmedqa,tsatsaronis-etal-2015-bioasq,jin-etal-2020-medqa,pal-etal-2022-medmcqa,nye-etal-2018-corpus,lehman-etal-2019-inferring,deyoung-etal-2020-evidence,wadden-etal-2020-fact}. Moreover, open-ended conclusion generation is difficult to evaluate reliably because reference-based metrics are incomplete, and LLM judges can vary substantially in calibration \citep{maynez-etal-2020-faithfulness,zheng-etal-2023-judging,liu-etal-2023-g,shi-etal-2024-judgingjudges,huang-etal-2025-empirical}.

To address these limitations, we present \textbf{\sys{}}, a large-scale \textbf{5.7M} dataset for biomedical conclusion generation from PubMed structured abstracts. Each instance pairs the non-conclusion sections of an abstract with its original author-written conclusion, yielding naturally occurring supervision for evidence-to-conclusion reasoning. In addition, \sys{} includes journal-level metadata, including \emph{biomedical category} labels and \emph{SJR} records. This combination of large-scale author-written supervision, broad biomedical coverage, and journal metadata enables analyses that are difficult to conduct in prior conclusion-generation settings. We further provide an initial empirical study using diverse LLMs, contrasting conclusion prompting against summary prompting, evaluating with a hybrid rule-based reference metrics and LLM judges, and examining robustness across judge backbones.

In summary, our work makes three contributions. First, we curate \textbf{\sys{}}, a \textbf{5.7M}-example dataset of PubMed structured abstracts for biomedical conclusion generation. Second, we augment the dataset with journal-level metadata, enabling aggregate and subgroup analysis across biomedical domains and venue strata. Third, we provide a first empirical study of the dataset by evaluating diverse LLMs, contrasting conclusion versus summary prompting, and studying the sensitivity of automatic evaluation to judge identity. We hope \sys{} serves as a reusable data resource for future study of scientific evidence-to-conclusion reasoning.

\section{Related work}

\begin{table}[t]
\centering
\small
\setlength{\tabcolsep}{4pt}
\renewcommand{\arraystretch}{1.15}
\begin{tabular}{p{3.0cm}p{1.35cm}ccccccc}
\toprule
& \multicolumn{5}{c}{\textbf{Dataset}} & \multicolumn{2}{c}{\textbf{Study Design}} \\
\cmidrule(lr){2-6}\cmidrule(lr){7-8}
\textbf{Name}
& \makecell{\textbf{Data}\\\textbf{Size}}
& \makecell{\textbf{Broad}\\\textbf{Biomed.}}
& \makecell{\textbf{Struc.}\\\textbf{Abs.}}
& \makecell{\textbf{Gold}\\\textbf{References}}
& \makecell{\textbf{Journal}\\\textbf{Metadata}}
& \makecell{\textbf{Summary}\\\textbf{Contrast}}
& \makecell{\textbf{Judge}\\\textbf{Robust.}} \\
\midrule

\citet{gao-etal-2024-evaluating} & 17.4K & \cmark & \cmark & \cmark & \xmark & \xmark & \xmark \\
\citet{shieh-etal-2019-towards}  & 195.7K & \xmark & \cmark & \cmark & \xmark & \xmark & \xmark \\
\citet{tang-etal-2022-echogen}   & 57.1K & \xmark & \xmark & \cmark & \xmark & \xmark & \xmark \\
\citet{tang-etal-2023-aligning}  & 200.2K & \xmark & \xmark & \cmark & \xmark & \xmark & \xmark \\
\citet{bastan-etal-2022-sume}    & 633K & \cmark & \xmark & \cmark & \xmark & \xmark & \xmark \\
\midrule
\rowcolor{black!3}
\textbf{\sys{}} & \textbf{5.7M} & \cmark & \cmark & \cmark & \cmark & \cmark & \cmark \\
\bottomrule
\end{tabular}
\caption{Comparison of \sys{} with conclusion-centric prior work. Checks denote properties that are central and explicitly emphasized by each resource or study. For multi-part resources, total dataset size sums all released components used in the paper.}
\label{tab:positioning_prior_work}
\end{table}

\paragraph{Adjacent biomedical reasoning resources.}

A large body of biomedical NLP work studies reasoning over scientific and clinical text, but not specifically the task of inferring an abstract conclusion from preceding evidence. PubMedQA and BioASQ evaluate biomedical question answering \citep{jin-etal-2019-pubmedqa,tsatsaronis-etal-2015-bioasq}; MedQA and MedMCQA focus on exam-style medical reasoning \citep{jin-etal-2020-medqa,pal-etal-2022-medmcqa}; EBM-NLP and Evidence Inference study treatment-effect reasoning from structured evidence \citep{nye-etal-2018-corpus,lehman-etal-2019-inferring,deyoung-etal-2020-evidence}; and SciFact evaluates scientific claim verification \citep{wadden-etal-2020-fact}. More recently, EvidenceBench studies sentence-level evidence extraction for biomedical hypotheses from full papers rather than conclusion generation from structured abstracts \citep{wang-etal-2025-evidencebench}. Structured abstracts and discourse-aware scientific summarization have also been widely studied \citep{teufel-moens-2002-summarizing,cohan-etal-2018-discourse,dernoncourt-lee-2017-pubmed,cachola-etal-2020-tldr,yasunaga-etal-2019-scisummnet}. \sys{} differs from these resources by centering benchmarking the evidence-to-conclusion reasoning step itself.

\vspace{-1em}

\paragraph{Conclusion-centric generation and reconstruction.}
The closest prior work to \sys{} studies conclusion reconstruction directly.~\citet{gao-etal-2024-evaluating} reconstruct conclusions of structured scientific abstracts to evaluate premise--conclusion alignment, treating conclusion generation mainly as an evaluation proxy.~\citet{shieh-etal-2019-towards} study conclusion generation for randomized controlled trial abstracts, and~\citet{tang-etal-2022-echogen} focus on echocardiography notes.~\citet{tang-etal-2023-aligning} study factual consistency in conclusion-oriented clinical-study summarization, while~\citet{bastan-etal-2022-sume} use large-scale PubMed conclusion generation as a training objective. In contrast, \sys{} contributes a broader \textbf{5.7M}-example PubMed resource with author-written targets and journal-level metadata, and uses it to study conclusion generation as a reasoning task rather than only as a proxy objective. Table~\ref{tab:positioning_prior_work} shows a detailed comparison.

\vspace{-1em}

\paragraph{Evaluation of open-ended scientific reasoning.}
Evaluating conclusion generation is challenging because open-ended outputs can differ in wording, scope, and detail while remaining partially valid. Prior work in summarization has shown that lexical overlap and fluency metrics are incomplete proxies for factual correctness~\citep{lin-2004-rouge,papineni-etal-2002-bleu,reimers-gurevych-2019-sentence,zhang-etal-2020-bertscore,maynez-etal-2020-faithfulness,kryscinski-etal-2020-evaluating,durmus-etal-2020-feqa,scialom-etal-2021-questeval,laban-etal-2022-summac}. LLM-as-a-judge provides a scalable alternative~\citep{zheng-etal-2023-judging,liu-etal-2023-g}, but recent work documents sensitivity to judge identity and grading scale, verbosity bias, position bias, and broader generalization concerns~\citep{dubois-etal-2024-lengthcontrolled,li2026grading,ye-etal-2024-justiceprejudice,huang-etal-2025-empirical,gu-etal-2024-survey-llmasajudge,zhu-etal-2025-judgelm}. These findings motivate the evaluation protocol used in this paper, but they are secondary to our main goal of introducing \sys{} as a large-scale dataset for biomedical conclusion generation.

\begin{figure}[!hbt]
\centering
\includegraphics[width=0.9\linewidth, page=1,trim=0cm 0.1cm 0 0,clip]{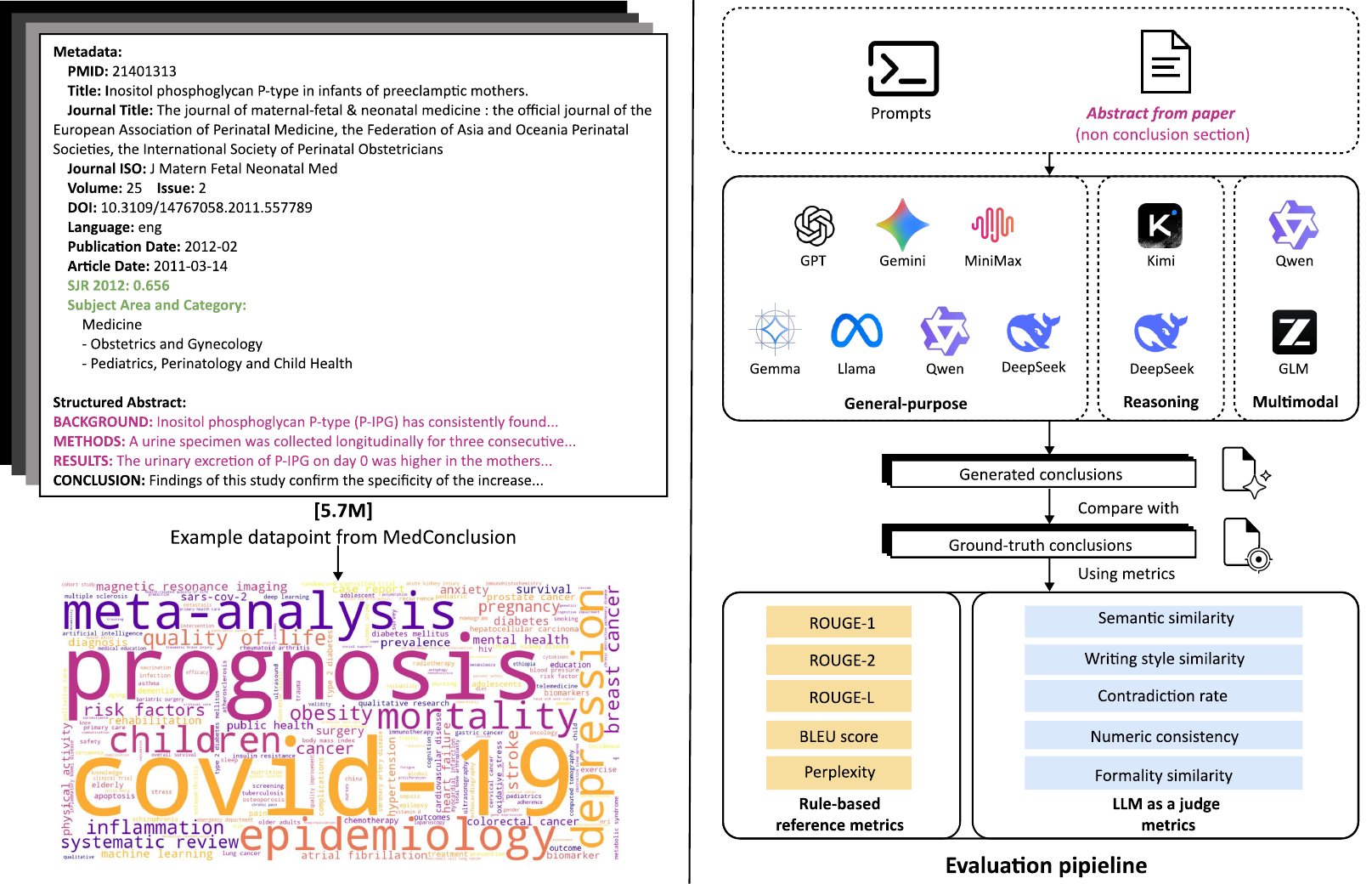}
\caption{Overview of \sys{} and the evaluation pipeline. Left: an example \sys{} instance, including article metadata, subject categories, and a structured abstract, where the non-conclusion sections are used as model input and the author-written conclusion serves as the gold reference. Right: the non-conclusion abstract is paired with prompts and given to diverse LLM families to generate conclusions, which are then compared against the ground-truth conclusion using both rule-based reference metrics and multi-dimensional LLM-as-a-judge metrics.}
\label{fig:pipeline}
\end{figure}

\section{Methodology}

\subsection{Data curation}

\textbf{\sys{}} is constructed from PubMed articles with structured abstracts published between 2000 and 2025. We identify candidate papers using PubMed's constraint \texttt{hasstructuredabstract} and collect the corresponding records for downstream processing.

\subsubsection{Data collection pipeline}
Our data collection pipeline is implemented with Entrez Direct (EDirect) \footnote{Kans J. Entrez® Direct: E-utilities on the Unix Command Line. 2013 Apr 23 [Updated 2025 Mar 25]. In: Entrez® Programming Utilities Help [Internet]. Bethesda (MD): National Center for Biotechnology Information (US); 2010-. Available from: https://www.ncbi.nlm.nih.gov/books/NBK179288/} and a custom XML parser. We first query PubMed for all UIDs satisfying \texttt{hasstructuredabstract} within the target time span, and deduplicate the retrieved identifiers. We then batch-download the corresponding PubMed XML records via the \texttt{epost} and \texttt{efetch} commands provided by EDirect, and parse each record into a JSONL representation containing article metadata, keywords, and structured abstract segments represented as (\texttt{label}, \texttt{nlm\_category}, \texttt{text}) tuples. Records with missing abstract labels are filtered out during parsing.

After parsing, we perform record-level deduplication using PMID, DOI, and normalized title. We then apply a rule-based cleaning procedure that keeps only English-language records with non-empty core bibliographic fields, normalizes date fields and missing metadata, and retains only articles with at least three abstract segments and at least one conclusion section. Conclusion sections are identified by matching normalized labels against a curated set of conclusion variants (Appendix~\ref{app:conclusion-label-variants}). We further remove records with malformed labels and clean keywords by dropping empty, overlong, or non-ASCII entries. The resulting cleaned corpus contains 5,692,839 structured abstract records with at least one conclusion section.

\subsubsection{Construction of \sys{} and dataset statistics}

The structured abstract records in \sys{} span a total of 3,772 unique journals. For each journal, we retrieved its subject category assignments and annual SJR scores from the SCImago Journal \& Country Rank (SJR) database,\footnote{\url{https://www.scimagojr.com}}
a publicly available bibliometric resource derived from Scopus.
Across the full corpus, the 3,772 journals are distributed across 141 subject categories. For SJR scores, we collected annual values from each journal's first indexed year in the SJR database through 2024. Dataset statistics and a formal definition of the SJR score are provided in
Appendix~\ref{app:MedConclusion-dataset-statistics}.



\subsection{Task, evaluation, and experimental setup}

Given a structured abstract with its \textbf{Conclusion} removed, let $x$ be the
concatenation of all remaining sections and let $y^\star$ be the original conclusion.
A model generates $\hat{y}=f_\theta(x)$.


We evaluate four prompting modes. \protect\circled{A} (default) asks the model to write a formal academic \textit{conclusion}, with no explicit length or style constraints. \protect\circled{B} asks the model to write a formal academic \textit{summary}, with no explicit length or style constraints. \protect\circled{C} asks the model to write a formal academic \textit{conclusion}, with explicit sentence- and word-count targets and instructions to match the abstract's writing style. \protect\circled{D} asks the model to write a formal academic \textit{summary}, with the same sentence- and word-count targets and the same instruction to match the abstract's writing style. Exact prompts are given in Appendix~\ref{sec:conclusion_summary_prompts}.

We score outputs with two classes of automatic metrics. First, we use multi-dimensional LLM-as-a-judge scoring. Given $(y^\star,\hat{y})$, the judge outputs five scores in $[0,100]$: \textbf{semantic similarity}, \textbf{writing style similarity}, \textbf{non-contradiction}, \textbf{numeric consistency}, and \textbf{formality similarity}. 
Second, we report lightweight diagnostics and reference-based metrics: word-count ratio, sentence-count ratio, embedding cosine similarity~\citep{reimers-gurevych-2019-sentence}, ROUGE-1/2/L~\citep{lin-2004-rouge}, BLEU~\citep{papineni-etal-2002-bleu}, and perplexity on the original and generated conclusions under a fixed external language model (\texttt{GPT-2}). Details for reference-based metrics could be found in Appendix~\ref{sec:reference_metrics}.

We evaluate a diverse set of LLMs spanning closed-source frontier models, open-source
instruction-tuned models, multimodal models, reasoning-oriented models, and small models.
For each model and prompting mode, we generate one output per instance using the
corresponding prompt template. All modes enforce a \emph{no-new-claims} instruction, and we
record length-ratio diagnostics to quantify format compliance. Due to cost constraints, we evaluate a randomly sampled 30K subset. Detailed model configurations are in Appendix~\ref{sec:models}.

To study evaluation sensitivity, we use \texttt{GPT-5.4-mini}~\citep{openai_gpt54_2026} as the primary judge and
\texttt{Gemini 3 Flash}~\citep{gemini3flash_modelcard_2025} as a secondary judge (prompts in Appendix~\ref{sec:judge_prompt}). Our two main comparisons are
\textbf{conclusion prompting versus summary prompting} and \textbf{judge-backbone robustness}.
The first tests whether models treat conclusion writing as a distinct discourse function
\citep{teufel-moens-2002-summarizing,cohan-etal-2018-discourse}; the second measures how
much absolute scores and model rankings depend on judge identity
\citep{zheng-etal-2023-judging,dubois-etal-2024-lengthcontrolled,shi-etal-2024-judgingjudges,ye-etal-2024-justiceprejudice,huang-etal-2025-empirical,gu-etal-2024-survey-llmasajudge}.

Figure~\ref{fig:pipeline} shows our overall pipeline, and Appendix~\ref{sec:example_data} shows an example data point.

\begin{table}[ht]
\centering
\small
\setlength{\tabcolsep}{4pt}
\begin{tabular}{lccccc}
\toprule
Model & \shortstack{Semantic\\Sim. $\uparrow$} & \shortstack{Writing Style\\Sim. $\uparrow$} & \shortstack{Non-Contradiction\\Rate $\uparrow$} & \shortstack{Numeric\\Consistency $\uparrow$} & \shortstack{Formality\\Sim. $\uparrow$} \\
\midrule
\multicolumn{6}{c}{\textbf{General-purpose Models}} \\
\midrule
GPT-5.4 & \textbf{73.22} & \textbf{71.21} & \textbf{84.61} & \textbf{88.24} & \textbf{89.80} \\
Gemini 3.1 Pro & \underline{71.87} & \underline{70.13} & \underline{82.02} & \underline{86.92} & \underline{89.49} \\
Gemini 3 Flash & 71.33 & 69.87 & 81.76 & 86.45 & 89.17 \\
Gemma-3-27B & 71.03 & 69.18 & 81.55 & 84.13 & 89.36 \\
DeepSeek-V3.2 & 69.47 & 68.21 & 80.31 & 86.22 & 88.59 \\
Llama-3.1-8B & 70.53 & 66.69 & 80.24 & 79.82 & 88.03 \\
MiniMax-M2.1 & 71.21 & 66.95 & 81.89 & 73.65 & 88.83 \\
Gemma-2-9B & 69.31 & 67.42 & 79.12 & 75.05 & 88.41 \\
Qwen3-4B & 69.80 & 66.35 & 78.96 & 71.78 & 88.47 \\
Qwen2.5-7B & 66.87 & 65.74 & 77.50 & 77.31 & 86.60 \\
Llama-3.2-1B & 54.17 & 50.69 & 66.14 & 82.69 & 78.35 \\
\midrule
\multicolumn{6}{c}{\textbf{Reasoning Models}} \\
\midrule
Kimi-K2 & 69.79 & 66.36 & 80.92 & 61.62 & 88.62 \\
DeepSeek-R1 & 68.93 & 48.06 & 79.67 & 75.58 & 75.91 \\
\midrule
\multicolumn{6}{c}{\textbf{Vision-Language Models}} \\
\midrule
GLM-4.6V & 70.86 & 68.83 & 80.50 & 80.19 & 88.87 \\
Qwen2.5-VL-7B & 68.96 & 64.74 & 78.73 & 71.82 & 87.34 \\
\bottomrule
\end{tabular}
\caption{LLM-as-Judge evaluation scores for conclusion generation. Models are grouped by primary capability. \textbf{Bold} and \underline{underline} denote the best and second-best scores, respectively.}
\label{tab:llm_judge_scores_metrics}
\end{table}

\begin{table}[ht]
\centering
\small
\setlength{\tabcolsep}{3pt}
\begin{tabular}{lccccccccc}
\toprule
Model & \shortstack{WC\\Ratio $\downarrow$} & \shortstack{SC\\Ratio $\downarrow$} & \shortstack{Embed.\\Sim. $\uparrow$} & \shortstack{ROUGE-\\1 $\uparrow$} & \shortstack{ROUGE-\\2 $\uparrow$} & \shortstack{ROUGE-\\L $\uparrow$} & BLEU $\uparrow$ & \shortstack{PPL\\Orig. $\downarrow$} & \shortstack{PPL\\Gen. $\downarrow$} \\
\midrule
\multicolumn{10}{c}{\textbf{General-purpose Models}} \\
\midrule
GPT-5.4 & 2.19 & 1.71 & \underline{0.77} & \underline{0.34} & \underline{0.10} & 0.21 & \underline{0.04} & 70.26 & 40.39 \\
Gemini 3.1 Pro & 2.28 & 1.75 & 0.73 & 0.33 & \underline{0.10} & 0.21 & \underline{0.04} & 70.26 & 34.54 \\
Gemini 3 Flash & 2.12 & 1.69 & 0.74 & \underline{0.34} & \underline{0.10} & 0.21 & \underline{0.04} & 70.26 & 33.87 \\
Gemma-3-27B & 2.49 & 2.01 & \underline{0.77} & 0.32 & 0.09 & 0.20 & \underline{0.04} & 70.26 & 30.46 \\
DeepSeek-V3.2 & \textbf{1.73} & \underline{1.38} & 0.76 & \textbf{0.35} & \textbf{0.11} & \textbf{0.23} & \textbf{0.05} & 70.26 & 47.14 \\
Llama-3.1-8B & 2.67 & 2.06 & 0.74 & 0.32 & \underline{0.10} & 0.20 & \underline{0.04} & 70.26 & \textbf{21.47} \\
MiniMax-M2.1 & 3.11 & 2.39 & 0.76 & 0.30 & 0.09 & 0.19 & 0.03 & 70.26 & 30.69 \\
Gemma-2-9B & 2.38 & 2.12 & \textbf{0.78} & 0.33 & \underline{0.10} & 0.21 & 0.04 & 70.26 & 30.06 \\
Qwen3-4B & 3.05 & 2.20 & 0.75 & 0.30 & 0.09 & 0.19 & 0.03 & 70.26 & 25.92 \\
Qwen2.5-7B & \underline{1.78} & 1.59 & 0.75 & \underline{0.34} & \textbf{0.11} & \underline{0.22} & \textbf{0.05} & 70.26 & 35.26 \\
Llama-3.2-1B & 1.82 & \textbf{1.23} & 0.72 & 0.31 & 0.09 & 0.20 & \underline{0.04} & 70.26 & 29.88 \\
\midrule
\multicolumn{10}{c}{\textbf{Reasoning Models}} \\
\midrule
Kimi-K2 & 2.90 & 2.79 & 0.75 & 0.30 & 0.09 & 0.18 & 0.03 & 70.26 & 60.76 \\
DeepSeek-R1 & 9.45 & 11.17 & 0.40 & 0.15 & 0.05 & 0.10 & 0.01 & 70.26 & 36.67 \\
\midrule
\multicolumn{10}{c}{\textbf{Vision-Language Models}} \\
\midrule
GLM-4.6V & 2.23 & 1.99 & 0.76 & \underline{0.34} & \textbf{0.11} & \underline{0.22} & \textbf{0.05} & 70.26 & 30.72 \\
Qwen2.5-VL-7B & 2.89 & 2.49 & 0.75 & 0.31 & \underline{0.10} & 0.20 & \underline{0.04} & 70.26 & \underline{23.52} \\
\bottomrule
\end{tabular}%
\caption{Rule-based evaluation scores for conclusion generation, same order as Table~\ref{tab:llm_judge_scores_metrics}.}
\label{tab:rule_based_scores_metrics}
\end{table}

\section{Results}

\subsection{Overall performance under conclusion generation}

Table~\ref{tab:llm_judge_scores_metrics} shows that \texttt{GPT-5.4} is the strongest model under the primary judge, leading all five judge dimensions. At the same time, models such as \texttt{Gemini 3.1 Pro}, \texttt{Gemini 3 Flash}, \texttt{DeepSeek-V3.2}, \texttt{Gemma-3-27B}, and \texttt{GLM-4.6V} lie within only a few points of the top model on most judge dimensions. This score compression suggests that current reference-comparison evaluation separates strong models only weakly, even though the task is not trivial.

Table~\ref{tab:rule_based_scores_metrics} tells a partly different story. \texttt{DeepSeek-V3.2} attains the best ROUGE-1/2/L and ties for the best BLEU, while \texttt{GPT-5.4} remains the strongest model under judge-based semantic and non-contradiction scores. Likewise, \texttt{Gemma-2-9B} has the highest embedding similarity despite clearly lower judge scores than the best closed models. Perplexity-based fluency diagnostics also decouple from task quality: several mid-sized open or multimodal models have substantially lower generated-text perplexity than \texttt{GPT-5.4}, yet they do not approach its semantic or contradiction scores. These mismatches indicate that lexical overlap, embedding similarity, fluency, and judge agreement capture different aspects of biomedical conclusion generation, and indicate our hybrid evaluation approach provides a more comprehensive evaluation than traditional reference-based metrics.

\begin{table}[ht]
\centering
\small
\setlength{\tabcolsep}{3pt}
\begin{tabular}{clccccc}
\toprule
Mode & Model & \shortstack{Semantic\\Sim. $\uparrow$} & \shortstack{Writing Style\\Sim. $\uparrow$} & \shortstack{Non-Contradiction\\Rate $\uparrow$} & \shortstack{Numeric\\Consistency $\uparrow$} & \shortstack{Formality\\Sim. $\uparrow$} \\
\midrule
\multicolumn{7}{c}{\textbf{Prompt without Formatting Restriction}} \\
\midrule
\multirow{2}{*}{\protect\circled{A}} & GPT-5.4        & 73.22 & 71.20 & 84.61 & 88.24 & 89.80 \\
                            & Gemini 3 Flash & 71.33 & 69.87 & 81.76 & 86.45 & 89.17 \\
\midrule
\multirow{2}{*}{\protect\circled{B}}    & GPT-5.4        & 72.11 & 62.60 & 83.96 & 66.24 & 88.47 \\
                            & Gemini 3 Flash & 71.04 & 61.55 & 82.44 & 58.09 & 88.11 \\
\midrule
\multicolumn{7}{c}{\textbf{Prompt with Length and Writing Style Restriction}} \\
\midrule
\multirow{2}{*}{\protect\circled{C}} & GPT-5.4        & 70.90 & 69.07 & 82.17 & 91.36 & 87.54 \\
                            & Gemini 3 Flash & 68.59 & 67.24 & 78.47 & 91.82 & 86.51 \\
\midrule
\multirow{2}{*}{\protect\circled{D}}    & GPT-5.4        & 64.99 & 60.13 & 78.76 & 74.06 & 85.35 \\
                            & Gemini 3 Flash & 64.16 & 62.29 & 75.51 & 77.66 & 85.28 \\
\bottomrule
\end{tabular}
\caption{LLM-as-Judge evaluation scores across prompt settings and generation modes. \protect\circled{A} and \protect\circled{C} are generating conclusions, whereas \protect\circled{B} and \protect\circled{D} are generating summaries.}
\label{tab:llm_judge_modes}
\end{table}

\subsection{Conclusion generation is not summary writing}
\label{sec:conc_vs_summ} 
Table~\ref{tab:llm_judge_modes} shows a clear discourse-function effect. Relative to \protect\circled{A}, \protect\circled{C} slightly reduces semantic and style similarity for both \texttt{GPT-5.4} and \texttt{Gemini 3 Flash}, but it improves numeric consistency to about 91 for both models. This suggests that explicit style and length control helps models better match the reference conclusion's level of numeric selectivity, even when it slightly constrains broader content choice.

When the target is changed from a conclusion to a summary, performance shifts more substantially. Under \protect\circled{D}, semantic similarity drops by about 7--8 points for both models relative to \protect\circled{A}, with even larger drops in writing style similarity and numeric consistency. \protect\circled{B} reveals an even more interesting pattern: semantic similarity rebounds to nearly the \protect\circled{A} level (within 1.11 points for \texttt{GPT-5.4} and 0.29 points for \texttt{Gemini 3 Flash}), but writing style similarity remains more than 8 points lower and numeric consistency collapses by 22.00 and 28.36 points, respectively. This suggests that unconstrained summaries often preserve the broad meaning of the abstract while selecting different details, especially numbers, scope qualifiers, and level of detail, than the published conclusion.

We therefore interpret these results as evidence that conclusion writing is behaviorally distinct from summary writing in this benchmark setting. Because the judge only compares outputs to the gold conclusion, the effect should be read as a difference in \emph{reference agreement and discourse targeting}, not as proof that summary-mode outputs are unsupported by the input. Some of the additional detail produced in summary modes may still be compatible with the source abstract even when it lowers the agreement with the reference conclusion. We further examine whether this distinction holds across biomedical subfields in Appendix~\ref{sec:category_conc_summ_gap}.

\begin{table}[ht]
\centering
\small
\setlength{\tabcolsep}{3pt}
\begin{tabular}{lccccc}
\toprule
Model & \shortstack{Semantic\\Sim. $\uparrow$} & \shortstack{Writing Style\\Sim. $\uparrow$} & \shortstack{Non-Contradiction\\Rate $\uparrow$} & \shortstack{Numeric\\Consistency $\uparrow$} & \shortstack{Formality\\Sim. $\uparrow$} \\
\midrule
\multicolumn{6}{c}{\textbf{Judge: GPT-5.4-mini}} \\
\midrule
GPT-5.4        & 73.22 & 71.20 & 84.61 & 88.24 & 89.80 \\
Gemini 3.1 Pro & 71.87 & 70.13 & 82.02 & 86.92 & 89.49 \\
Gemini 3 Flash & 71.33 & 69.87 & 81.76 & 86.45 & 89.17 \\
\midrule
\multicolumn{6}{c}{\textbf{Judge: Gemini 3 Flash}} \\
\midrule
GPT-5.4        & 84.30 & 71.49 & 97.51 & 98.18 & 92.50 \\
Gemini 3.1 Pro & 82.64 & 68.41 & 96.58 & 97.53 & 90.70 \\
Gemini 3 Flash & 82.59 & 70.04 & 96.62 & 97.28 & 91.46 \\
\bottomrule
\end{tabular}
\caption{LLM-as-Judge evaluation scores: GPT-5.4-mini vs Gemini 3 Flash as judge.}
\label{tab:llm_judge_gptvsgemini}
\end{table}

\subsection{Judge robustness: score scale shifts across judges}

Table~\ref{tab:llm_judge_gptvsgemini} shows large absolute calibration shifts when the judge backbone is changed. Switching from \texttt{GPT-5.4-mini} to \texttt{Gemini 3 Flash} as judge raises semantic similarity, non-contradiction, and numeric consistency across the same three generation models. By contrast, writing style similarity changes only modestly. Thus, the absolute score scale is highly judge-dependent. However, ranking is relatively stable. \texttt{GPT-5.4} remains the top generator under both judges across all five dimensions, yet the middle ordering can flip occasionally.

\section{Analysis}




We further analyze benchmark behavior across journal-level subgroups to understand
where conclusion generation is relatively easier or harder.
In particular, we study variation along two axes: \textbf{journal prestige}, measured
by SJR score, and \textbf{biomedical category}.
These analyses are descriptive rather than part of the benchmark definition itself,
and are intended to reveal whether performance differences are associated with venue
prestige or with the broader structure of biomedical subfields.
All results in this section use \texttt{GPT-5.4} under setting~\protect\circled{A}
as the representative condition unless otherwise noted.




\subsection{SJR score}

\begin{figure}[!hbt]
\centering
\includegraphics[width=\linewidth]{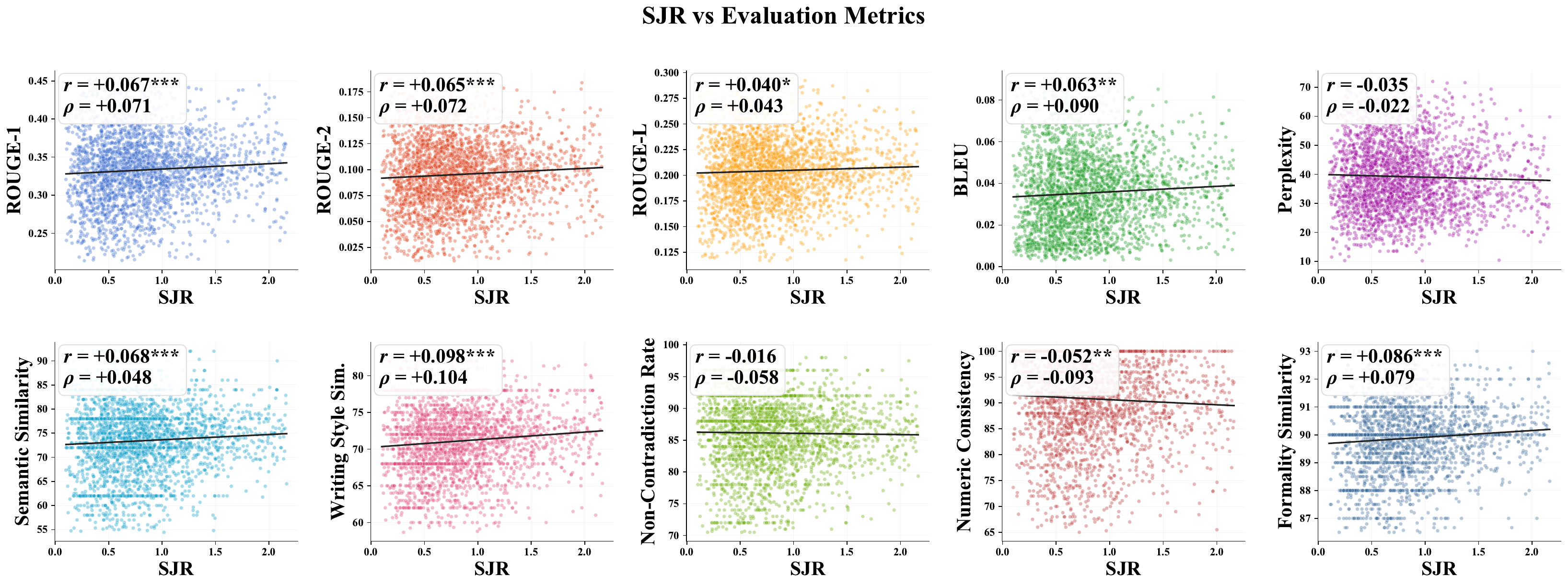}
\caption{Scatter plots of journal-level SJR scores versus evaluation metrics for \texttt{GPT-5.4}
under the \protect\circled{A} setting (outliers removed). Each point represents one journal,
aggregated by mean score. The top row shows reference-based metrics and the bottom row shows LLM-judge dimensions.
Pearson ($r$) and Spearman ($\rho$) correlations are annotated in each panel
(${*}p < 0.05$, ${**}p < 0.01$, ${***}p < 0.001$).}
\label{fig:sjr_vs_metrics_scatter}
\end{figure}

Figure~\ref{fig:sjr_vs_metrics_scatter} plots journal-level SJR scores against each evaluation metric for \texttt{GPT-5.4} under setting \protect\circled{A}.
Most reference-based and judge-based metrics show small but statistically significant positive associations with SJR: ROUGE-1, ROUGE-2, Semantic Similarity, Writing Style Similarity, and Formality Similarity are all significant at $p < 0.001$.
By contrast, Perplexity and Non-Contradiction Rate show no significant trend, while Numeric Consistency exhibits a small but significant negative correlation.
These results suggest that journals with higher SJR scores tend to produce abstracts whose conclusions are slightly easier to match in terms of lexical overlap and writing style, yet this advantage does not extend to factual consistency dimensions. The overall effect sizes remain modest, indicating that venue prestige is a weak rather than dominant predictor of conclusion-generation difficulty.

\subsection{Category}

\begin{figure}[!hbt]
\centering

\begin{subfigure}{0.9\linewidth}
    \centering
    \includegraphics[width=\linewidth,page=1,trim=0 0cm 0 0,clip]{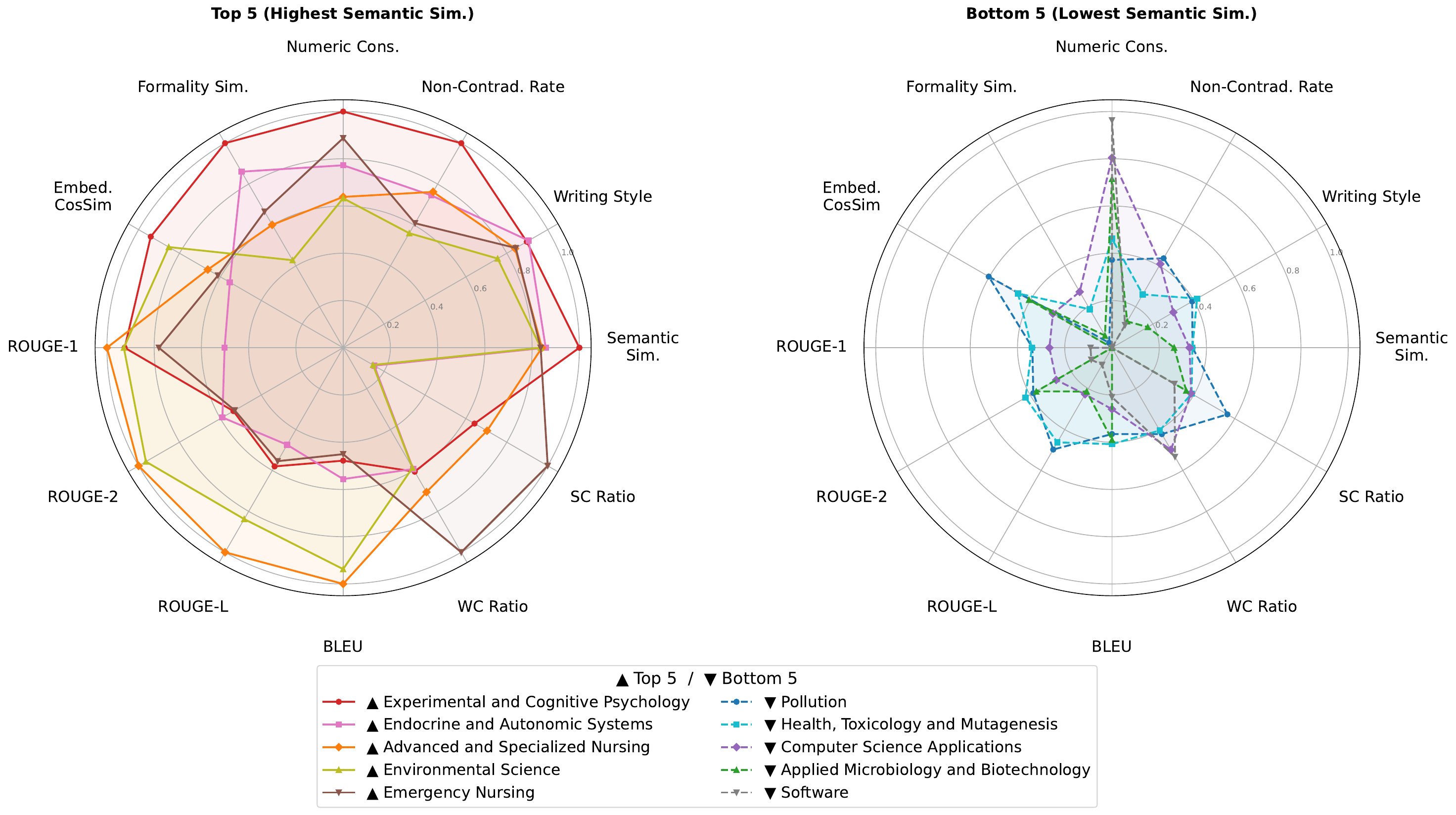}
    \caption{Top 5 and bottom 5 biomedical categories ranked by mean Semantic Similarity.}
\end{subfigure}

\vspace{0em}

\begin{subfigure}{0.9\linewidth}
    \centering
    \includegraphics[width=\linewidth,page=1,trim=0 0cm 0 0,clip]{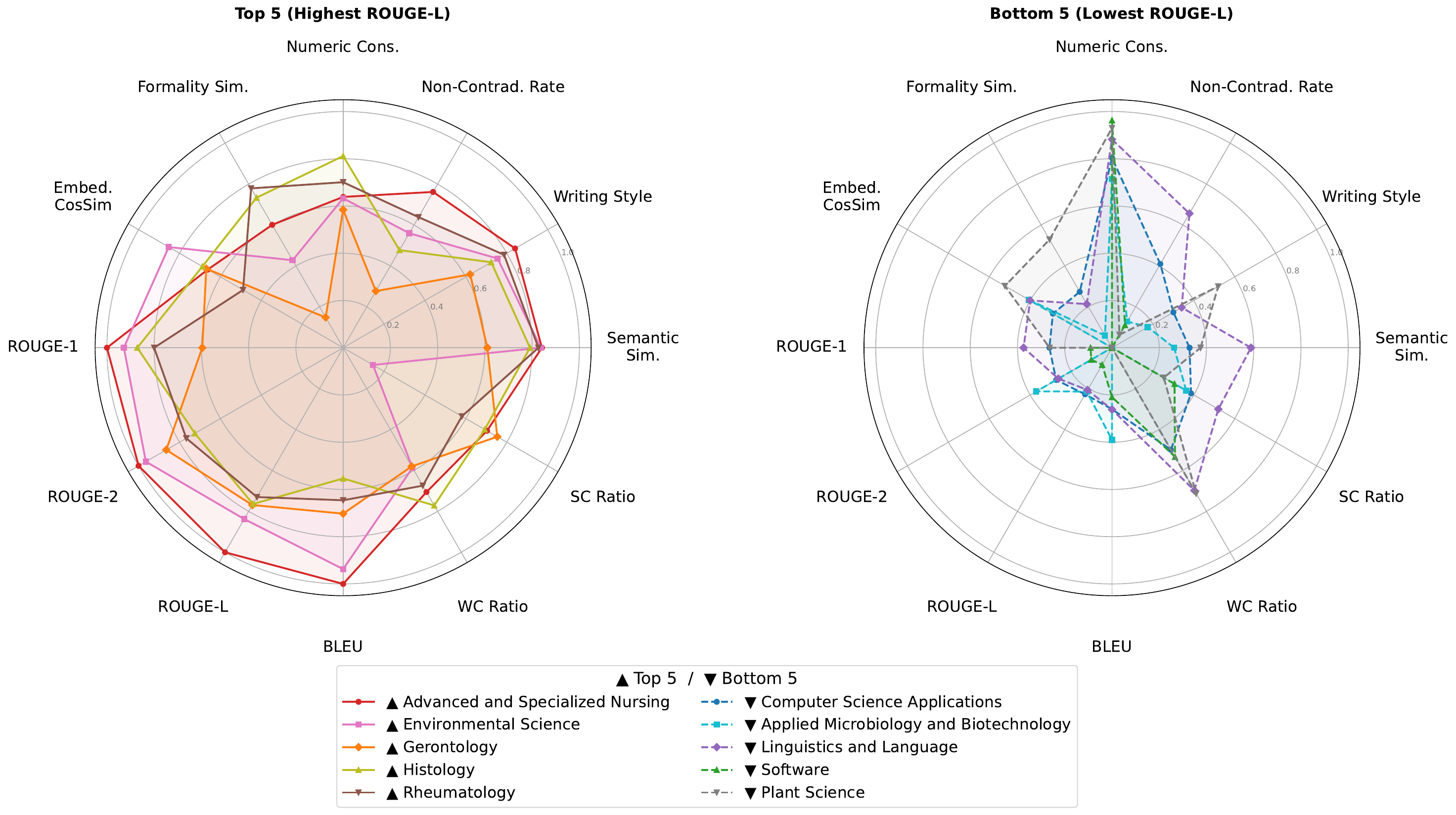}
    \caption{Top 5 and bottom 5 biomedical categories ranked by mean ROUGE-L.}
\end{subfigure}

\caption{Radar chart comparison of the top 5 and bottom 5 biomedical categories ranked by both mean Semantic Similarity and mean ROUGE-L for \texttt{GPT-5.4} under setting \protect\circled{A}. Each axis is normalized to $[0, 1]$.}
\label{fig:radar_category}
\end{figure}

Figure~\ref{fig:radar_category} compares the top-5 and bottom-5 biomedical categories under two ranking criteria: mean semantic similarity (panel~a) and mean ROUGE-L (panel~b), both for \texttt{GPT-5.4} under setting~\protect\circled{A}, with all twelve metrics min-max normalized to $[0,1]$.

When categories are ranked by semantic similarity (panel~a), the top-5 categories form large, uniformly filled polygons: high semantic similarity co-occurs with high writing style similarity, numeric consistency, non-contradiction rate, and lexical overlap metrics alike. In contrast, when categories are ranked by ROUGE-L (panel~b), the top-5 polygons are visibly lopsided. ROUGE-1/2/L axes are high by construction, but judge-based dimensions such as writing style and numeric consistency do not follow. For example, Gerontology ranks among the top-5 by ROUGE-L yet falls well below the semantic-similarity top-5 on writing style and numeric consistency.

This asymmetry reveals that \textbf{lexical overlap with the reference conclusion is not a reliable proxy for overall conclusion quality}. Categories whose generated conclusions share many surface $n$-grams with the gold reference do not necessarily match its rhetorical style, numeric selectivity, or broader discourse structure. By contrast, high semantic similarity appears to act as a more holistic quality indicator that correlates with strong performance across both reference-based and judge-based dimensions. This observation reinforces the motivation for our hybrid evaluation protocol (Section~\ref{sec:conc_vs_summ}): relying on ROUGE or BLEU alone would mask meaningful quality differences that only LLM-as-a-judge scoring can detect.

The bottom-5 categories under both rankings share substantial overlap: Software, Computer Science Applications, and Applied Microbiology and Biotechnology appear in both lists, confirming that these interdisciplinary, non-clinical fields are consistently the hardest for conclusion generation regardless of the evaluation axis. Their radar profiles are highly irregular. For instance, Software has the lowest semantic similarity across all 112 categories (61.0) yet among the highest numeric consistency (96.4), illustrating that no single metric suffices to characterize conclusion-generation difficulty. Appendix~\ref{app:category_analysis} provides representative generation examples from both high- and low-performing categories.

\section{Conclusion}

We introduce \sys{}, a large-scale benchmark of 5.7M structured PubMed abstracts for biomedical conclusion generation, paired with author-written conclusions and journal-level metadata. Our experiments show that conclusion generation is behaviorally distinct from summary writing, that strong LLMs remain closely clustered under current automatic metrics, and that absolute LLM-judge scores are sensitive to judge identity. Analyses across journal prestige and biomedical categories further show that task difficulty is heterogeneous and that lexical-overlap metrics alone do not adequately capture conclusion quality. 


\newpage

\bibliography{colm2026_conference}
\bibliographystyle{colm2026_conference}

\newpage

\appendix
\section{Example data}\label{sec:example_data}

\begin{table}[!htb]
\begin{tcolorbox}[title=Example Datapoint from \sys{},
  colback=white, colframe=black!60]
\textbf{Metadata:}\\
\hspace*{1em}\textbf{PMID:} \texttt{21401313}\\
\hspace*{1em}\textbf{Title:} Inositol phosphoglycan P-type in infants of preeclamptic mothers.\\
\hspace*{1em}\textbf{Journal Title:} The journal of maternal-fetal \& neonatal medicine : the official journal of the European Association of Perinatal Medicine, the Federation of Asia and Oceania Perinatal Societies, the International Society of Perinatal Obstetricians\\
\hspace*{1em}\textbf{Journal ISO:} \textit{J Matern Fetal Neonatal Med}\\
\hspace*{1em}\textbf{Volume:} 25 \hspace{1.5em}
\textbf{Issue:} 2\\
\hspace*{1em}\textbf{DOI:} \texttt{10.3109/14767058.2011.557789}\\
\hspace*{1em}\textbf{Language:} eng\\
\hspace*{1em}\textbf{Publication Date:} 2012-02\\
\hspace*{1em}\textbf{Article Date:} 2011-03-14\\
\hspace*{1em}\textbf{SJR 2012:} 0.656\\
\hspace*{1em}\textbf{Subject Area and Category:}\\
\hspace*{2em}Medicine\\
\hspace*{2em}- Obstetrics and Gynecology\\
\hspace*{2em}- Pediatrics, Perinatology and Child Health\medskip\\
\textbf{Structured Abstract:}\\
\inputinstr{\textbf{BACKGROUND:} Inositol phosphoglycan P-type (P-IPG) has consistently found to be elevated during active preeclampsia, although the biosynthetic source has to be identified yet. This multicenter prospective cross-sectional case-control study evaluated the fetus/newborn as the source of P-IPG.}\medskip\\
\inputinstr{\textbf{METHODS:} A urine specimen was collected longitudinally for three consecutive days after delivery from 90 newborns and their mothers, and ordered according to clinical diagnosis of preeclampsia, gestational hypertension, or healthy pregnancy.}\medskip\\
\inputinstr{\textbf{RESULTS:} The urinary excretion of P-IPG on day 0 was higher in the mothers in all groups (\textit{p} $<$ 0.05) with higher levels in preeclamptic women (\textit{p} $<$ 0.01) in the mothers compared to their newborns in the preeclamptic group (\textit{p} $<$ 0.01). The difference persisted at least two days post partum.}\medskip\\
\groundtruthinstr{\textbf{CONCLUSION:} Findings of this study confirm the specificity of the increase in urinary excretion of P-IPG in preeclamptic mothers at day of birth compared to healthy pregnancy and GH, but does not extend to their newborns.}
\end{tcolorbox}
\caption{Example datapoint used for prompt construction and evaluation. In the structured abstract, the non-conclusion sections are highlighted separately from the \texttt{CONCLUSION} section to indicate that the former are used as model input, while the conclusion serves as the ground-truth reference for evaluation.}
\label{tab:example-datapoint}
\end{table}

\newpage
\section{Prompts for conclusion/summary generation}\label{sec:conclusion_summary_prompts}


\subsection[Prompts for conclusion generation (A and B)]
{Prompts for conclusion generation (\protect\circled{A} and \protect\circled{C})}

\begin{table}[!htb]
\begin{tcolorbox}[title=Conclusion Generation (\protect\circled{A} vs.\ \protect\circled{C}),
  colback=white, colframe=blue!80]
\textbf{System:} You are a senior scientist in generating conclusions for scientific papers.\medskip\\
\textbf{User:}\\
You will be provided with a structured abstract of a scientific paper.\\
The abstract contains sections such as Background, Objective, Methods, Results, etc., but the corresponding Conclusion section is missing.\medskip\\
Your task is to infer and write the most plausible \textbf{CONCLUSION} section that would appear in this abstract.\medskip\\
Here are the requirements:\\
\hspace*{1em}- Output ONLY the text for the conclusion itself. Do NOT include any section headers or explanations.\\
\hspace*{1em}- Only use the information provided by the abstract to derive your conclusion.\\
\hspace*{1em}- Do NOT introduce new experiments, datasets, numerical values, or claims that are not supported by the abstract.\medskip\\
\writinginstr{Please generate the conclusion in the same writing style as the given abstract sections. The conclusion should be a concise paragraph with \texttt{\detokenize{{sen_num}}} sentences totalling \texttt{\detokenize{{word_num}}} words.}\medskip\\
\norestrictioninstr{Use formal academic writing style.}\medskip\\
Structured Abstract:\\
\texttt{\detokenize{<Abstract>}}\\
\texttt{\detokenize{{abstract_text}}}\\
\texttt{\detokenize{</Abstract>}}
\end{tcolorbox}
\caption{Prompts for conclusion generation. The highlighted variants distinguish the constrained writing setting (\protect\circled{C}), which enforces sentence and word count targets and asks the model to match the abstract's writing style, from the unconstrained setting (\protect\circled{A}), which instead only requires a formal academic style without length constraints.}
\label{tab:conclusion-grouped}
\end{table}

\newpage
\subsection[Prompts for summary generation (B and D)]
{Prompts for summary generation (\protect\circled{B} and \protect\circled{D})}

\begin{table}[!htb]
\begin{tcolorbox}[title=Summary Generation (\protect\circled{B} vs.\ \protect\circled{D}),
  colback=white, colframe=blue!80]
\textbf{System:} You are a senior scientist in generating summaries for scientific papers.\medskip\\
\textbf{User:}\\
You will be provided with a structured abstract of a scientific paper.\\
The abstract contains sections such as Background, Objective, Methods, Results, etc.\medskip\\
Your task is to summarize the core information in the given abstract sections.\medskip\\
Here are the requirements:\\
\hspace*{1em}- Output ONLY the text for the summary itself. Do NOT include any section headers or explanations.\\
\hspace*{1em}- Only use the information provided by the abstract to derive your summary.\\
\hspace*{1em}- Do NOT introduce new experiments, datasets, numerical values, or claims that are not supported by the abstract.\medskip\\
\writinginstr{Please generate the summary in the same writing style as the given abstract sections. The summary should be a concise paragraph with \texttt{\detokenize{{sen_num}}} sentences totalling \texttt{\detokenize{{word_num}}} words.}\medskip\\
\norestrictioninstr{Use formal academic writing style.}\medskip\\
Structured Abstract:\\
\texttt{\detokenize{<Abstract>}}\\
\texttt{\detokenize{{abstract_text}}}\\
\texttt{\detokenize{</Abstract>}}
\end{tcolorbox}
\caption{Prompts for summary generation. The highlighted variants distinguish the constrained writing setting (\protect\circled{D}), which requires the summary to follow the abstract's writing style and satisfy sentence and word count targets, from the unconstrained setting (\protect\circled{B}), which only specifies a formal academic style and leaves length unrestricted.}
\label{tab:summary-grouped}
\end{table}

\newpage

\section{Prompts for LLM judges}\label{sec:judge_prompt}

\begin{table}[!htb]
\begin{tcolorbox}[title=LLM Judge Prompt,
  colback=white, colframe=red!80]
\textbf{System:} You are an expert evaluator of scientific writing.\medskip\\
\textbf{User:}\\
Your task is to compare the Generated Conclusion against the Original (Reference) Conclusion and score multiple dimensions from 0 to 100 (decimals allowed). Use ONLY the two conclusions provided. Do NOT provide any explanations.\medskip\\
Scoring dimensions:\\
\hspace*{1em}- semantic similarity: How similar the meaning and core claims are.\\
\hspace*{1em}- writing style similarity: How similar the tone, phrasing, structure, and rhetorical style are.\\
\hspace*{1em}- contradiction rate: Degree of contradiction between the Generated Conclusion and the Original Conclusion.\\
\hspace*{2em}- 100 = no contradiction\\
\hspace*{2em}- 0 = severe contradiction\\
\hspace*{1em}- numeric consistency: Consistency of all numerical information, quantities, directions, and magnitudes.\\
\hspace*{2em}- 100 = fully consistent or no numeric content in either text\\
\hspace*{2em}- 0 = major numeric inconsistency\\
\hspace*{1em}- formality similarity: Similarity in academic/formal writing level and register.\medskip\\
INPUT:\\
\hspace*{1em}- Original Conclusion (reference): \texttt{\detokenize{{original_conclusion}}}\\
\hspace*{1em}- Generated Conclusion: \texttt{\detokenize{{generated_conclusion}}}\medskip\\
OUTPUT FORMAT (STRICT):\\
\texttt{\detokenize{{"semantic similarity": <0-100>, "writing style similarity": <0-100>, "contradiction rate": <0-100>, "numeric consistency": <0-100>, "formality similarity": <0-100>}}}
\end{tcolorbox}
\caption{Prompt for LLM judge}
\label{tab:llm-judge-prompt}
\end{table}



\newpage
\section{Reference-based metrics}\label{sec:reference_metrics}

In addition to LLM-as-a-judge scoring, we report a set of lightweight diagnostics and
reference-based metrics that compare the generated conclusion $\hat{y}$ against the
author-written reference conclusion $y^\star$. These metrics are inexpensive to compute,
easy to reproduce, and provide complementary signals about lexical overlap, semantic
proximity, length control, and fluency. We do not treat any single metric as a complete
measure of conclusion quality; rather, we use them as a bundle of auxiliary indicators.

Let $|y|_{\mathrm{word}}$ denote the number of words in a text $y$, and let
$|y|_{\mathrm{sent}}$ denote its number of sentences.

\paragraph{Word-count ratio.}
To measure length matching at the tokenized word level, we compute
\begin{equation}
\mathrm{WCR}(\hat{y}, y^\star)
=
\frac{|\hat{y}|_{\mathrm{word}}}{|y^\star|_{\mathrm{word}}}.
\end{equation}
A value close to $1$ indicates that the generated conclusion has similar length to the
reference. Values below $1$ indicate shorter generations, while values above $1$ indicate
longer generations.

\paragraph{Sentence-count ratio.}
To assess structural length control at the sentence level, we compute
\begin{equation}
\mathrm{SCR}(\hat{y}, y^\star)
=
\frac{|\hat{y}|_{\mathrm{sent}}}{|y^\star|_{\mathrm{sent}}}.
\end{equation}

\paragraph{Embedding cosine similarity.}
To capture semantic similarity beyond surface lexical overlap, we encode $\hat{y}$ and
$y^\star$ using an off-the-shelf \textsc{all-mpnet-base-v2}\footnote{\url{https://huggingface.co/sentence-transformers/all-mpnet-base-v2}} sentence embedding model and compute the cosine similarity between the
two vector representations~\citep{reimers-gurevych-2019-sentence}:
\begin{equation}
\mathrm{CosSim}(\hat{y}, y^\star)
=
\frac{\phi(\hat{y})^\top \phi(y^\star)}
{\|\phi(\hat{y})\| \, \|\phi(y^\star)\|},
\end{equation}
where $\phi(\cdot)$ denotes the embedding function. Higher values indicate greater
semantic proximity.

\paragraph{ROUGE.}
We report ROUGE-1, ROUGE-2, and ROUGE-L~\citep{lin-2004-rouge}. ROUGE-1 and ROUGE-2 measure
unigram and bigram overlap, respectively, while ROUGE-L measures longest-common-subsequence
overlap. These metrics quantify the extent to which the generated conclusion reuses words
and short phrases appearing in the reference. Because multiple valid conclusions may use
different wording, ROUGE should be interpreted as a lexical-overlap signal rather than a
direct measure of scientific correctness.

\paragraph{BLEU.}
We also report BLEU~\citep{papineni-etal-2002-bleu}, which measures $n$-gram precision of
the generated text with a brevity penalty. As with ROUGE, BLEU is sensitive to phrasing and
therefore is best viewed as an approximate indicator of closeness to the reference wording,
not as a standalone measure of conclusion quality.

\paragraph{Perplexity under an external language model.}
To estimate fluency and distributional typicality, we compute perplexity using a fixed
external language model, \texttt{GPT-2}, on both the reference conclusion $y^\star$ and the
generated conclusion $\hat{y}$. For a sequence of tokens
$y=(w_1,\dots,w_T)$, perplexity under language model $p$ is
\begin{equation}
\mathrm{PPL}(y)
=
\exp\!\left(
-\frac{1}{T}\sum_{t=1}^{T}\log p(w_t \mid w_{<t})
\right).
\end{equation}
Lower perplexity indicates that the text is more probable under the external language
model. Reporting perplexity for both $y^\star$ and $\hat{y}$ helps contextualize whether
model-generated conclusions are comparably fluent to author-written ones under the same
scoring model.

\paragraph{Implementation notes.}
All reference-based metrics are computed between each generated conclusion $\hat{y}$ and
its paired author-written conclusion $y^\star$. We aggregate scores over the evaluation
set using the arithmetic mean unless otherwise noted. Since these metrics emphasize
different aspects of generation quality, we recommend interpreting them jointly with the
LLM-as-a-judge results in the main paper.

\newpage
\section{Additional category analysis}\label{app:category_analysis}


\subsection{Example 1}

\begin{table}[!htb]
\begin{tcolorbox}[title={Example: Experimental and Cognitive Psychology (HIGH)},
  colback=white, colframe=black!60]
\textbf{Metadata:}\\
\hspace*{1em}\textbf{PMID:} \texttt{32422422}\\
\hspace*{1em}\textbf{Journal Title:} Cortex\\
\hspace*{1em}\textbf{Publication Year:} 2020\\
\hspace*{1em}\textbf{SJR 2020:} 1.786\\
\hspace*{1em}\textbf{Subject Area and Category:}\\
\hspace*{2em}- Experimental and Cognitive Psychology\\
\textbf{Structured Abstract:}\\
\inputinstr{\textbf{BACKGROUND:} Numerous studies have shown visuoperceptual/visuospatial deficits in dementia with Lewy bodies (DLB) and Alzheimer's disease (AD). Visual texture recognition is also impaired in patients with DLB and AD. Although patients with DLB often exhibit visual misidentifications of objects, there are few studies on the relationships between visual texture recognition and viewpoints for object recognition.}\\
\inputinstr{\textbf{OBJECTIVES:} The aim of this study was to clarify how viewpoints, textures, and visual cognitive functions affect object recognition and result in visual misidentifications in patients with DLB or AD.}\\
\inputinstr{\textbf{METHODS:} A total of 37 patients with probable DLB and 58 with probable AD and 32 age-matched healthy controls underwent neuropsychological and visuoperceptual assessments, and performed object identification tasks under four conditions (non-canonical view + blurry texture, non-canonical view + clear texture, canonical view + blurry texture, canonical view + clear texture). The relationship between object identification and other visuoperceptual functions was analyzed.}\\
\inputinstr{\textbf{RESULTS:} Patients with DLB and AD exhibited significantly impaired object recognition under non-canonical viewing with blurry texture conditions, with the DLB patients exhibiting a significantly worse performance than the AD patients. Patients with DLB and AD exhibited visual misidentifications during object identification tasks under non-canonical viewing. In patients with DLB, the number of visual misidentifications was significantly correlated with the scores of visual texture recognition.}\\
\groundtruthinstr{\textbf{CONCLUSION:} The present study showed that significantly impaired object recognition in patients with DLB under the influences by both viewpoint and visual texture and in those with AD under the influence by viewpoint. Visual misidentification in object recognition could be associated with impaired visual texture recognition in DLB.}\\
\generatedinstr{\textbf{GENERATED:} Object recognition is particularly vulnerable in DLB and, to a lesser extent, in AD when objects are presented from non-canonical viewpoints with degraded texture information. The greater impairment in DLB and the association between visual misidentifications and visual texture recognition in this group suggest that deficits in processing texture cues, especially under atypical viewing conditions, contribute to visual misidentification. These findings indicate that altered viewpoint and texture processing are important mechanisms underlying visual recognition disturbances in DLB.}
\end{tcolorbox}
\caption{Example from \textit{Experimental and Cognitive Psychology}. ROUGE-1: 0.480, ROUGE-2: 0.130, ROUGE-L: 0.272, BLEU: 0.068, Perplexity: 41.1, Semantic Sim.: 88, Writing Style Sim.: 84, Non-Contradiction Rate: 96, Numeric Consistency: 100, Formality Sim.: 92.}
\label{tab:example-experimental-and-cognitive-psychology}
\end{table}


\newpage
\subsection{Example 2}

\begin{table}[!htb]
\begin{tcolorbox}[title={Example: Endocrine and Autonomic Systems (HIGH)},
  colback=white, colframe=black!60]
\textbf{Metadata:}\\
\hspace*{1em}\textbf{PMID:} \texttt{37580720}\\
\hspace*{1em}\textbf{Journal Title:} Thyroid research\\
\hspace*{1em}\textbf{Publication Year:} 2023\\
\hspace*{1em}\textbf{SJR 2023:} 0.492\\
\hspace*{1em}\textbf{Subject Area and Category:}\\
\hspace*{2em}- Endocrine and Autonomic Systems\\
\textbf{Structured Abstract:}\\
\inputinstr{\textbf{PURPOSE:} Lacrimal gland enlargement can be a feature of thyroid eye disease (TED). Unilateral or asymmetric lacrimal gland enlargement is poorly described and may impede diagnosis. We present the histological and clinical findings of four patients with asymmetric lacrimal gland enlargement.}\\
\inputinstr{\textbf{METHODS:} A retrospective case note review was performed for patients over two tertiary orbital clinics (Royal Adelaide Hospital, South Australia and the Sussex Eye Hospital, Brighton, United Kingdom) presenting with an asymmetrical lacrimal gland enlargement with a background of TED that underwent biopsy to exclude alternate diagnoses. Baseline data was collected for each patient and histopathological images and reports were reviewed.}\\
\inputinstr{\textbf{RESULTS:} All four patients were hyperthyroid at time of lacrimal gland biopsy. Biopsy demonstrated nonspecific, lymphoid aggregates, typically of B cell type, with no diagnostic findings to support lymphocyte clonality or IgG4-related disease. One biopsy specimen demonstrated evidence of some fibrosis.}
\\
\groundtruthinstr{\textbf{CONCLUSION:} Asymmetrical lacrimal gland enlargement can occur as part of the TED spectrum but may require biopsy to exclude alternate pathology. Histology demonstrates a non-specific lymphocytic infiltrate.}\\
\generatedinstr{\textbf{GENERATED:} In patients with thyroid eye disease, asymmetric lacrimal gland enlargement may occur and can mimic alternative orbital pathology. In this series, biopsy showed only nonspecific benign lymphoid infiltration without evidence of lymphoma or IgG4-related disease, supporting asymmetric lacrimal gland involvement as a possible manifestation of TED. These findings may aid clinical recognition, although biopsy remains important when the presentation is atypical and alternative diagnoses must be excluded.}
\end{tcolorbox}
\caption{Example from \textit{Endocrine and Autonomic Systems}. ROUGE-1: 0.358, ROUGE-2: 0.065, ROUGE-L: 0.232, BLEU: 0.020, Perplexity: 46.5, Semantic Sim.: 88, Writing Style Sim.: 82, Non-Contradiction Rate: 96, Numeric Consistency: 100, Formality Sim.: 94.}
\label{tab:example-endocrine-and-autonomic-systems}
\end{table}


\newpage
\subsection{Example 3}

\begin{table}[!htb]
\begin{tcolorbox}[title={Example: Advanced and Specialized Nursing (HIGH)},
  colback=white, colframe=black!60]
\textbf{Metadata:}\\
\hspace*{1em}\textbf{PMID:} \texttt{32951753}\\
\hspace*{1em}\textbf{Journal Title:} Complementary therapies in medicine\\
\hspace*{1em}\textbf{Publication Year:} 2020\\
\hspace*{1em}\textbf{SJR 2020:} 0.58\\
\hspace*{1em}\textbf{Subject Area and Category:}\\
\hspace*{2em}- Advanced and Specialized Nursing\\
\textbf{Structured Abstract:}\\
\inputinstr{\textbf{BACKGROUND AND OBJECTIVE:} Walnut intake is considered a healthy dietary approach worldwide, particularly as a nutritional tool for the management of obesity and cardiometabolic disorders. Among these lines, leptin and adiponectin, as well as glycemic biomarkers, deserve further attention. We aimed to examine the impact of walnut intake on circulation levels of leptin and adiponectin through a systematic review and meta-analysis of randomized clinical trials (RCTs); secondarily, assessing the glycemic profile as well.}\medskip\\
\inputinstr{\textbf{METHODS:} The literature search was implemented in four following databases: Web of Science, Scopus, PubMed/Medline, and Google Scholar, thus, determining studies that measured the effects of walnut consumption on adiponectin, leptin, and glycemic biomarkers levels from 2004 up to December 2019.}\medskip\\
\inputinstr{\textbf{RESULTS:} Fourteen trials were include in the meta-analysis, with an intervention period ranging from 5 weeks to 12 months.Walnut intake increased leptin (weighted mean difference (WMD): 2.502 ng/mL; 95 \% CI: 2.147-2.856, p  0.001) and adiponectin (WMD: 0.440 ng/mL; 95 \% CI: 0.323 to 0.557, p  0.001) levels. Pertaining to glycemic biomarkers, neither overall analyses nor sub-analyses corroborated with changes in fasting blood glucose (WMD: 0.500 mg/dL, 95 \% CI: -0.596, 1.596, p = 0.371), insulin (WMD: -0.21 mg/dL, 95 \% CI: -0.67, 0.24, p = 0.367), and glycated hemoglobin (WMD: 0.004 mg/dL, 95 \% CI: -0.041, 0.049, p = 0.870) concentrations.}\\
\groundtruthinstr{\textbf{CONCLUSION:} Walnut intake may increase leptin and adiponectin levels but does not improve glycemic biomarkers.}\medskip\\
\generatedinstr{\textbf{GENERATED:} Walnut consumption appears to favorably modulate adipokine status by increasing circulating adiponectin and leptin levels, while showing no significant effect on fasting blood glucose, insulin, or glycated hemoglobin. These findings suggest that walnuts may contribute to obesity and cardiometabolic management primarily through effects on adipokine regulation rather than glycemic control. Further well-designed randomized clinical trials are warranted to confirm these effects and clarify their clinical relevance.}
\end{tcolorbox}
\caption{Example from \textit{Advanced and Specialized Nursing}. ROUGE-1: 0.198, ROUGE-2: 0.000, ROUGE-L: 0.123, BLEU: 0.004, Perplexity: 15.7, Semantic Sim.: 88, Writing Style Sim.: 82, Non-Contradiction Rate: 96, Numeric Consistency: 100, Formality Sim.: 90.}
\label{tab:example-advanced-and-specialized-nursing}
\end{table}

\newpage
\subsection{Example 4}

\begin{table}[!htb]
\begin{tcolorbox}[title={Example: Environmental Science (HIGH)},
  colback=white, colframe=black!60]
\textbf{Metadata:}\\
\hspace*{1em}\textbf{PMID:} \texttt{31054526}\\
\hspace*{1em}\textbf{Journal Title:} Environmental research\\
\hspace*{1em}\textbf{Publication Year:} 2019\\
\hspace*{1em}\textbf{SJR 2019:} 1.52\\
\hspace*{1em}\textbf{Subject Area and Category:}\\
\hspace*{2em}Environmental Science\\
\textbf{Structured Abstract:}\\
\inputinstr{\textbf{BACKGROUND:} Hypertension and air pollution are two important risk factors for cardiovascular morbidity and mortality. Although several studies suggest that air pollution has a significant impact on blood pressure, studies on long-term effects are sparse and still controversial.}\medskip\\
\inputinstr{\textbf{OBJECTIVE:} To evaluate the effects of exposure of outdoor workers to different levels of traffic-generated PM2.5 on blood pressure.}\\
\inputinstr{\textbf{DESIGN:} This is an observational panel study.}\\
\inputinstr{\textbf{PARTICIPANTS:} 88 non-smoking workers exposed to different concentrations of air pollution were evaluated weekly along four successive weeks.}\\
\inputinstr{\textbf{MEASUREMENTS:} In each week, personal monitoring of 24-h PM2.5 concentration and 24-h ambulatory blood pressure were measured. The association between blood pressure variables and PM2.5, adjusted for age, body mass index, time in job, daily work hours, diabetes, hypertension and cholesterol was assessed by means of multiple linear regression models fitted by least squares.}\medskip\\
\inputinstr{\textbf{RESULTS:} Exposure to PM2.5 (ranging from 8.5 to 89.7 $\mu$g/m[3]) is significantly and consistently associated with an increase in average blood pressure. An elevation of 10 $\mu$g/m[3] in the concentration of PM2.5 is associated with increments of 3.9 mm Hg (CI 95\% = [1.5; 6.3]) in average systolic 24-h blood pressure for hypertensive and/or diabetic workers.}
\medskip\\
\groundtruthinstr{\textbf{CONCLUSION:} Exposure to fine particles, predominantly from vehicular traffic, is associated with elevated blood pressure in hypertensive and/or diabetic workers.}\medskip\\
\generatedinstr{\textbf{GENERATED:} Personal exposure to traffic-generated PM2.5 was positively associated with higher 24-hour blood pressure in this cohort of outdoor workers, with particularly pronounced effects among workers with hypertension and/or diabetes. These findings support the view that long-term exposure to fine particulate air pollution contributes to elevated blood pressure and may increase cardiovascular risk in susceptible populations.}
\end{tcolorbox}
\caption{Example from \textit{Environmental Science}. ROUGE-1: 0.375, ROUGE-2: 0.231, ROUGE-L: 0.300, BLEU: 0.039, Perplexity: 15.8, Semantic Sim.: 88, Writing Style Sim.: 82, Non-Contradiction Rate: 98, Numeric Consistency: 100, Formality Sim.: 95.}
\label{tab:example-environmental-science}
\end{table}

\newpage
\subsection{Example 5}
\begin{table}[!htb]
\begin{tcolorbox}[title={Example: Emergency Nursing (HIGH)},
  colback=white, colframe=black!60]
\textbf{Metadata:}\\
\hspace*{1em}\textbf{PMID:} \texttt{21458134}\\
\hspace*{1em}\textbf{Journal Title:} Resuscitation\\
\hspace*{1em}\textbf{Publication Year:} 2011\\
\hspace*{1em}\textbf{SJR 2011:} 1.736\\
\hspace*{1em}\textbf{Subject Area and Category:}\\
\hspace*{2em}- Emergency Nursing\medskip\\
\textbf{Structured Abstract:}\\
\inputinstr{\textbf{AIM:} Body mass index (BMI) may influence the quality of cardiopulmonary resuscitation and may influence prognosis after cardiac arrest. To review the direct effect of obesity on outcome after cardiac arrest, the following cohort study was conducted.}\medskip\\
\inputinstr{\textbf{METHODS:} This study based on a cardiac arrest registry comprising all adult patients with cardiac arrest of non-traumatic origin and restoration of spontaneous circulation (ROSC) admitted to the department of emergency medicine of a tertiary-care facility. Data were collected between January 1992 and December 2007 according to the Utstein criteria. We assessed the association between BMI according to the WHO classification (underweight, BMI18.5; normal weight, 18.5-24.9; overweight, 25.0-29.9; obese $\geq$ 30), six-month survival and neurological recovery.}\medskip\\
\inputinstr{\textbf{RESULTS:} Analysis was carried out on a total of 1915 adult patients (32\% female). Patients had a median age of 59 years (interquartile range [IQR] 49-70) and a median BMI of 26.0 (IQR 23.9-29.1). Survival to six months was 50\%. There was no significant difference in survival between the BMI groups (underweight 46\%, normal weight 47\%, overweight 52\%, obese 51\%). In a multivariate analysis neurological outcome was better in overweight patients as compared to subjects with normal BMI (odds ratio 1.35; 95\% confidence interval 1.02-1.79).}
\medskip\\
\groundtruthinstr{\textbf{CONCLUSION:} Body mass index may have no direct influence on six-month survival after cardiac arrest, but patients with moderately elevated BMI may have a better neurological prognosis.}\medskip\\
\generatedinstr{\textbf{GENERATED:} In this cohort of adult patients with non-traumatic cardiac arrest and ROSC, BMI was not significantly associated with six-month survival. However, overweight patients showed a modestly better neurological outcome than patients with normal BMI after adjustment for confounders, suggesting that excess body weight does not adversely affect prognosis after cardiac arrest and may be associated with improved neurological recovery.}
\end{tcolorbox}
\caption{Example from \textit{Emergency Nursing}. ROUGE-1: 0.341, ROUGE-2: 0.140, ROUGE-L: 0.182, BLEU: 0.027, Perplexity: 20.2, Semantic Sim.: 88, Writing Style Sim.: 82, Non-Contradiction Rate: 94, Numeric Consistency: 96, Formality Sim.: 90.}
\label{tab:example-emergency-nursing}
\end{table}

\newpage
\subsection{Example 6}
\begin{table}[!htb]
\begin{tcolorbox}[title={Example: Pollution (LOW)},
  colback=white, colframe=black!60]
\textbf{Metadata:}\\
\hspace*{1em}\textbf{PMID:} \texttt{36497716}\\
\hspace*{1em}\textbf{Journal Title:} International journal of environmental research and public health\\
\hspace*{1em}\textbf{Publication Year:} 2022\\
\hspace*{1em}\textbf{SJR 2022:} 0.828\\
\hspace*{1em}\textbf{Subject Area and Category:}\\
\hspace*{2em}- Pollution\\\medskip\\
\textbf{Structured Abstract:}\\
\inputinstr{\textbf{BACKGROUND:} Urinary incontinence (UI) and poor sleep negatively affect health-related quality of life (HRQoL). This study explored the UI-related factors and the relationships between UI, sleep quality, and HRQoL.}\medskip\\
\inputinstr{\textbf{METHODS:} This cross-sectional study collected data from 237 women with type 2 diabetes. Multivariate logistic regression was conducted to identify the factors associated with UI. One-way analysis of variance was used to compare the mean sleep quality and HRQoL scores of women without UI and those who experienced UI of varying severities. Correlation coefficients were estimated, and multivariate linear regression was conducted to examine the relationships between UI severity, sleep quality, and HRQoL.}\medskip\\
\inputinstr{\textbf{RESULTS:} Of the 237 women, 115 (48.52\%) experienced UI and 139 (58.65\%) were poor sleepers. The three factors associated with UI were advanced age, a higher body mass index, and a history of vaginal delivery. Significant associations between UI severity and sleep quality and between sleep quality and HRQoL were revealed. UI severity and night-time voiding frequency were both associated with sleep quality.}
\medskip\\
\groundtruthinstr{\textbf{CONCLUSION:} One factor associated with UI (body mass index) is modifiable. UI severity is associated with sleep quality as the possible influence of night-time voiding frequency on sleep quality has been considered.}\medskip\\
\generatedinstr{\textbf{GENERATED:} Urinary incontinence is common among women with type 2 diabetes and is associated with older age, higher body mass index, and a history of vaginal delivery. Greater UI severity, particularly with more frequent night-time voiding, is related to poorer sleep quality, and poor sleep is in turn associated with worse health-related quality of life. These findings highlight the importance of routine assessment and management of UI and sleep problems in this population to improve overall well-being.}
\end{tcolorbox}
\caption{Example from \textit{Pollution}. ROUGE-1: 0.378, ROUGE-2: 0.165, ROUGE-L: 0.252, BLEU: 0.030, Perplexity: 18.8, Semantic Sim.: 62, Writing Style Sim.: 68, Non-Contradiction Rate: 78, Numeric Consistency: 100, Formality Sim.: 85.}
\label{tab:example-pollution}
\end{table}

\newpage
\subsection{Example 7}
\begin{table}[!htb]
\begin{tcolorbox}[title={Example: Health, Toxicology and Mutagenesis (LOW)},
  colback=white, colframe=black!60]
\textbf{Metadata:}\\
\hspace*{1em}\textbf{PMID:} \texttt{28934092}\\
\hspace*{1em}\textbf{Journal Title:} Environmental health perspectives\\
\hspace*{1em}\textbf{Publication Year:} 2017\\
\hspace*{1em}\textbf{SJR 2017:} 3.41\\
\hspace*{1em}\textbf{Subject Area and Category:}\\
\hspace*{2em}- Health, Toxicology and Mutagenesis\\\medskip\\
\textbf{Structured Abstract:}\\
\inputinstr{\textbf{BACKGROUND:} Some epidemiologic and laboratory studies suggest that insecticides are related to increased breast cancer risk, but the evidence is inconsistent. Women engaged in agricultural work or who reside in agricultural areas may experience appreciable exposures to a wide range of insecticides.}\medskip\\
\inputinstr{\textbf{OBJECTIVE:} We examined associations between insecticide use and breast cancer incidence among wives of pesticide applicators (farmers) in the prospective Agricultural Health Study.}\medskip\\
\inputinstr{\textbf{METHODS:} Farmers and their wives provided information on insecticide use, demographics, and reproductive history at enrollment in 1993-1997 and in 5-y follow-up interviews. Cancer incidence was determined via cancer registries. Among 30,594 wives with no history of breast cancer before enrollment, we examined breast cancer risk in relation to the women's and their husbands' insecticide use using Cox proportional hazards regression to estimate adjusted hazard ratios (HRs) and 95\% confidence intervals (CIs).}\medskip\\
\inputinstr{\textbf{RESULTS:} During an average 14.7-y follow-up, 39\% of the women reported ever using insecticides, and 1,081 were diagnosed with breast cancer ... Among women who did not apply pesticides, the husband's use of fonofos was associated with elevated risk, although no exposure-response trend was observed.}
\medskip\\
\groundtruthinstr{\textbf{CONCLUSION:} Use of several organophosphate insecticides was associated with elevated breast cancer risk. However, associations for the women's and husbands' use of these insecticides showed limited concordance. Ongoing cohort follow-up may help clarify the relationship, if any, between individual insecticide exposures and breast cancer risk. https://doi.org/10.1289/EHP1295.}\medskip\\
\generatedinstr{\textbf{GENERATED:} Overall insecticide use was not associated with breast cancer incidence among wives of pesticide applicators; however, use of specific insecticides, particularly the organophosphates chlorpyrifos and terbufos, was associated with modestly increased risk. These associations appeared to be driven primarily by premenopausal breast cancer, with little evidence of variation by estrogen receptor status. The observed elevation in risk with husbands' fonofos use among nonapplying wives, in the absence of an exposure-response trend, warrants cautious interpretation. These findings suggest that certain insecticides may contribute to breast cancer risk in this agricultural population and merit further investigation.}
\end{tcolorbox}
\caption{Example from \textit{Health, Toxicology and Mutagenesis}. ROUGE-1: 0.381, ROUGE-2: 0.083, ROUGE-L: 0.245, BLEU: 0.027, Perplexity: 21.5, Semantic Sim.: 62, Writing Style Sim.: 78, Non-Contradiction Rate: 58, Numeric Consistency: 100, Formality Sim.: 92.}
\label{tab:example-health-toxicology-and-mutagenesis}
\end{table}

\newpage
\subsection{Example 8}
\begin{table}[!htb]
\begin{tcolorbox}[title={Example: Computer Science Applications (LOW)},
  colback=white, colframe=black!60]
\textbf{Metadata:}\\
\hspace*{1em}\textbf{PMID:} \texttt{39014177}\\
\hspace*{1em}\textbf{Journal Title:} International journal of computer assisted radiology and surgery\\
\hspace*{1em}\textbf{Publication Year:} 2025\\
\hspace*{1em}\textbf{SJR 2025:} 0.658\\
\hspace*{1em}\textbf{Subject Area and Category:}\\
\hspace*{2em}- Computer Science Applications\\\medskip\\
\textbf{Structured Abstract:}\\
\inputinstr{\textbf{PURPOSE:} Augmented reality guidance in laparoscopic liver resection requires the registration of a preoperative 3D model to the intraoperative 2D image. However, 3D-2D liver registration poses challenges owing to the liver's flexibility, particularly in the limited visibility conditions of laparoscopy. Although promising, the current registration methods are computationally expensive and often necessitate manual initialisation.}\medskip\\
\inputinstr{\textbf{METHODS:} The first neural model predicting the registration (NM) is proposed, represented as 3D model deformation coefficients, from image landmarks. The strategy consists in training a patient-specific model based on synthetic data generated automatically from the patient's preoperative model. A liver shape modelling technique, which further reduces time complexity, is also proposed.}\medskip\\
\inputinstr{\textbf{RESULTS:} The NM method was evaluated using the target registration error measure, showing an accuracy on par with existing methods, all based on numerical optimisation. Notably, NM runs much faster, offering the possibility of achieving real-time inference, a significant step ahead in this field.}
\medskip\\
\groundtruthinstr{\textbf{CONCLUSION:} The proposed method represents the first neural method for 3D-2D liver registration. Preliminary experimental findings show comparable performance to existing methods, with superior computational efficiency. These results suggest a potential to deeply impact liver registration techniques.}\medskip\\
\generatedinstr{\textbf{GENERATED:} A patient-specific neural approach can accurately estimate 3D-2D liver registration from image landmarks while substantially reducing computational cost relative to optimisation-based methods. By leveraging automatically generated synthetic data and an efficient liver shape modelling strategy, the proposed framework removes the need for costly numerical optimisation and supports the prospect of real-time augmented reality guidance in laparoscopic liver resection.}
\end{tcolorbox}
\caption{Example from \textit{Computer Science Applications}. ROUGE-1: 0.303, ROUGE-2: 0.082, ROUGE-L: 0.182, BLEU: 0.025, Perplexity: 96.5, Semantic Sim.: 62, Writing Style Sim.: 54, Non-Contradiction Rate: 78, Numeric Consistency: 100, Formality Sim.: 88.}
\label{tab:example-computer-science-applications}
\end{table}

\newpage
\subsection{Example 9}
\begin{table}[!htb]
\begin{tcolorbox}[title={Example: Applied Microbiology and Biotechnology (LOW)},
  colback=white, colframe=black!60]
\textbf{Metadata:}\\
\hspace*{1em}\textbf{PMID:} \texttt{26497155}\\
\hspace*{1em}\textbf{Journal Title:} Journal of applied microbiology\\
\hspace*{1em}\textbf{Publication Year:} 2016\\
\hspace*{1em}\textbf{SJR 2016:} 0.84\\
\hspace*{1em}\textbf{Subject Area and Category:}\\
\hspace*{2em}Biochemistry, Genetics and Molecular Biology\\
\hspace*{2em}Immunology and Microbiology\\
\hspace*{2em}Medicine\\
\hspace*{2em}- Applied Microbiology and Biotechnology\\
\hspace*{2em}- Biotechnology\\
\hspace*{2em}- Medicine\\
\textbf{Structured Abstract:}\\
\inputinstr{\textbf{AIMS:} The effect of ohmic heating (OH) in a pilot plant system which had a zig-zag shaped (elbow-type) ohmic heater with five sequential voltage electrodes was investigated on Bacillus cereus spores in a commercial tsuyu sauce.}\\
\inputinstr{\textbf{METHODS AND RESULTS:} The electrical field was fixed at 26$\cdot$7 V cm(-1) with an alternating current frequency of 25 kHz. Raw tsuyu sauce (50 l) inoculated with B. cereus spores was submitted in a 4 $\times$ 3 factorial design to the OH system and heated at 95, 105, 115, and 125$^{\circ}$C each for 30, 60, and 90 s. Survival of B. cereus spores and colour change in the commercial tsuyu sauce were both measured before and after treatment. As the treatment temperature and time increased, the number of surviving B. cereus spores decreased. The OH treatment in a bath-type process at 105$^{\circ}$C for more than 30 s resulted in the total inactivation of the inoculated B. cereus spores (average 5$\cdot$4 log reductions to undetectable levels after treatment). The OH protocol of heating at 105$^{\circ}$C for 60 s which ensure complete eradication of the inoculated spores without compromising product quality was chosen and investigated for its suitability for commercial application on bulk quantities of samples (500 l). Reliable and reproducible reductions in B. cereus spore counts of 4$\cdot$7-5$\cdot$5 log CFU ml(-1) (mean $\pm$ standard deviation = 5$\cdot$1 $\pm$ 0$\cdot$3 CFU ml(-1)) were achieved by the selected protocol of the continuous OH treatment (105$^{\circ}$C for 60 s).}\\
\inputinstr{\textbf{SIGNIFICANCE AND IMPACT OF THE STUDY:} This procedure will enhance the microbiological quality of liquid foods while minimizing quality deterioration.}
\\
\groundtruthinstr{\textbf{CONCLUSION:} This study suggests that OH treatment with five sequential elbow-type electrodes has great potential as an industrial sterilizing method for liquid food contaminated with B. cereus spores.}\\
\generatedinstr{\textbf{GENERATED:} Ohmic heating using the pilot-scale zig-zag electrode system effectively inactivated Bacillus cereus spores in commercial tsuyu sauce in a temperature- and time-dependent manner. Heating at 105$^{\circ}$C for 60 s achieved complete spore eradication in the bath-type process and produced reliable, reproducible reductions during continuous treatment of bulk samples, while maintaining product colour quality. These findings indicate that the developed ohmic heating protocol is a practical approach for commercial processing of liquid foods to improve microbiological safety with minimal quality deterioration.}
\end{tcolorbox}
\caption{Example from \textit{Applied Microbiology and Biotechnology}. ROUGE-1: 0.177, ROUGE-2: 0.036, ROUGE-L: 0.088, BLEU: 0.006, Perplexity: 79.3, Semantic Sim.: 62, Writing Style Sim.: 58, Non-Contradiction Rate: 78, Numeric Consistency: 12, Formality Sim.: 84.}
\label{tab:example-applied-microbiology-and-biotechnology}
\end{table}

\newpage
\subsection{Example 10}
\begin{table}[!htb]
\begin{tcolorbox}[title={Example: Software (LOW)},
  colback=white, colframe=black!60]
\textbf{Metadata:}\\
\hspace*{1em}\textbf{PMID:} \texttt{29157445}\\
\hspace*{1em}\textbf{Journal Title:} Computer methods and programs in biomedicine\\
\hspace*{1em}\textbf{Publication Year:} 2018\\
\hspace*{1em}\textbf{SJR 2018:} 0.753\\
\hspace*{1em}\textbf{Subject Area and Category:}\\
\hspace*{2em}- Software\\\\
\textbf{Structured Abstract:}\\
\inputinstr{\textbf{BACKGROUND AND OBJECTIVES:} Diabetic retinopathy (DR) is one of the leading causes of preventable blindness in the world. Its earliest sign are red lesions, a general term that groups both microaneurysms (MAs) and hemorrhages (HEs). In daily clinical practice, these lesions are manually detected by physicians using fundus photographs. However, this task is tedious and time consuming, and requires an intensive effort due to the small size of the lesions and their lack of contrast. Computer-assisted diagnosis of DR based on red lesion detection is being actively explored due to its improvement effects both in clinicians consistency and accuracy. Moreover, it provides comprehensive feedback that is easy to assess by the physicians. Several methods for detecting red lesions have been proposed in the literature, most of them based on characterizing lesion candidates using hand crafted features, and classifying them into true or false positive detections. Deep learning based approaches, by contrast, are scarce in this domain due to the high expense of annotating the lesions manually.}\\
\inputinstr{\textbf{METHODS:} In this paper we propose a novel method for red lesion detection based on combining both deep learned and domain knowledge. Features learned by a convolutional neural network (CNN) are augmented by incorporating hand crafted features. Such ensemble vector of descriptors is used afterwards to identify true lesion candidates using a Random Forest classifier.}\\
\inputinstr{\textbf{RESULTS:} We empirically observed that combining both sources of information significantly improve results with respect to using each approach separately. Furthermore, our method reported the highest performance on a per-lesion basis on DIARETDB1 and e-ophtha, and for screening and need for referral on MESSIDOR compared to a second human expert.}
\\
\groundtruthinstr{\textbf{CONCLUSION:} Results highlight the fact that integrating manually engineered approaches with deep learned features is relevant to improve results when the networks are trained from lesion-level annotated data. An open source implementation of our system is publicly available at https://github.com/ignaciorlando/red-lesion-detection.}\\
\generatedinstr{\textbf{GENERATED:} The proposed red lesion detection framework, which integrates CNN-derived features with hand-crafted descriptors and classifies lesion candidates using a Random Forest, provides superior performance to either deep learned or hand-crafted features alone. Its strong results across DIARETDB1, e-ophtha, and MESSIDOR indicate that combining data-driven representation learning with domain knowledge is an effective strategy for accurate red lesion detection and for supporting diabetic retinopathy screening and referral decisions in clinical practice.}
\end{tcolorbox}
\caption{Example from \textit{Software}. ROUGE-1: 0.264, ROUGE-2: 0.050, ROUGE-L: 0.198, BLEU: 0.008, Perplexity: 106.9, Semantic Sim.: 62, Writing Style Sim.: 58, Non-Contradiction Rate: 88, Numeric Consistency: 100, Formality Sim.: 84.}
\label{tab:example-software}
\end{table}

\newpage
\subsection{The conclusion--summary distinction holds across categories}
\label{sec:category_conc_summ_gap}

Section~\ref{sec:conc_vs_summ} shows that summary-mode outputs recover most of the semantic similarity of conclusion-mode outputs while diverging sharply in writing style and numeric consistency. We now test whether this pattern is universal or driven by a subset of categories, by computing the per-category gap (Mode~\protect\circled{A}$-$\protect\circled{B}) across all five judge dimensions for \texttt{GPT-5.4}.

Across all 112 categories, the gap is \emph{always} positive for writing style similarity (min $+3.8$, mean $+8.3$, max $+13.7$) and numeric consistency (min $+11.3$, mean $+21.6$, max $+41.3$).

Table~\ref{tab:category_gap_top10} shows the ten categories with the largest $|\Delta\text{Semantic Similarity}|$. Even within this group, the gap magnitudes vary substantially across dimensions: numeric consistency ranges from $+18.6$ to $+33.0$ and writing style from $+4.5$ to $+13.7$, indicating that the specific dimensions along which the two modes diverge are category-dependent. Biotechnology is a notable outlier: it is the only entry where summary-mode outputs are semantically \emph{closer} to the reference ($\Delta = -4.0$), yet numeric consistency still drops by $+27.2$ points.

Table~\ref{tab:category_gap_bottom10} isolates the ten categories where the semantic gap is smallest ($|\Delta| < 0.3$). Even here, writing style still differs by $+4.8$ to $+8.4$ points and numeric consistency by $+14.3$ to $+23.1$ points. Semantic convergence does not imply behavioral equivalence: summaries differ from conclusions in phrasing, structure, and numeric detail even when their meaning is indistinguishable. This confirms that the conclusion--summary distinction is a structural property of the two discourse functions, not an artifact of category-level heterogeneity.

\begin{table}[H]
\centering
\small
\setlength{\tabcolsep}{3pt}
\begin{tabular}{lccccc}
\toprule
Category
& \shortstack{$\Delta$ Semantic\\Sim.}
& \shortstack{$\Delta$ Writing\\Style}
& \shortstack{$\Delta$ Non-Contrad.\\Rate}
& \shortstack{$\Delta$ Numeric\\Cons.}
& \shortstack{$\Delta$ Formality\\Sim.} \\
\midrule
Immunology \& Microbiology          & $+6.0$ & $+13.0$ & $+4.8$ & $+28.0$ & $+2.3$ \\
Pharm., Toxicology \& Pharmaceutics & $+5.5$ & $+13.5$ & $+4.9$ & $+33.0$ & $+2.4$ \\
Exp.\ \& Cognitive Psychology       & $+5.3$ & $+7.9$  & $+3.4$ & $+27.5$ & $+2.1$ \\
Arts \& Humanities                   & $+5.1$ & $+12.3$ & $+6.5$ & $+32.4$ & $+2.8$ \\
Virology                             & $+4.8$ & $+13.0$ & $+7.2$ & $+23.2$ & $+3.1$ \\
Nursing                              & $+4.4$ & $+13.7$ & $+3.8$ & $+30.3$ & $+1.8$ \\
Social Psychology                    & $+4.1$ & $+11.4$ & $+2.9$ & $+18.6$ & $+2.1$ \\
Biotechnology                        & $-4.0$ & $+4.5$  & $-4.6$ & $+27.2$ & $+0.6$ \\
Education                            & $+3.9$ & $+12.3$ & $+3.6$ & $+27.6$ & $+1.7$ \\
Health Informatics                   & $+3.5$ & $+12.7$ & $+4.0$ & $+31.5$ & $+3.1$ \\
\bottomrule
\end{tabular}
\caption{Top 10 categories ranked by $|\Delta\text{Semantic Similarity}|$ (Mode~\protect\circled{A}$-$\protect\circled{B}; \texttt{GPT-5.4}). Gap magnitudes vary substantially across dimensions within this group, indicating that the specific dimensions along which conclusion and summary modes diverge are category-dependent.}
\label{tab:category_gap_top10}
\end{table}

\begin{table}[H]
\centering
\small
\setlength{\tabcolsep}{3pt}
\begin{tabular}{lccccc}
\toprule
Category
& \shortstack{$\Delta$ Semantic\\Sim.}
& \shortstack{$\Delta$ Writing\\Style}
& \shortstack{$\Delta$ Non-Contrad.\\Rate}
& \shortstack{$\Delta$ Numeric\\Cons.}
& \shortstack{$\Delta$ Formality\\Sim.} \\
\midrule
Cancer Research                   & $+0.2$ & $+7.6$ & $-0.5$ & $+16.7$ & $+0.7$ \\
Organic Chemistry                 & $-0.2$ & $+7.1$ & $-1.2$ & $+18.1$ & $+1.2$ \\
Endocrinology, Diabetes \& Metab. & $+0.1$ & $+7.3$ & $-0.2$ & $+14.3$ & $+0.9$ \\
Cellular \& Mol.\ Neuroscience    & $-0.1$ & $+6.8$ & $+0.4$ & $+16.5$ & $+1.1$ \\
Hepatology                        & $-0.1$ & $+4.8$ & $+0.7$ & $+15.1$ & $+1.1$ \\
Biochemistry                      & $-0.1$ & $+5.4$ & $+0.6$ & $+16.4$ & $+0.2$ \\
Applied Psychology                & $+0.0$ & $+5.9$ & $-1.2$ & $+23.1$ & $+2.5$ \\
Dentistry                         & $-0.0$ & $+8.4$ & $-0.5$ & $+19.6$ & $+1.1$ \\
Critical Care \& ICM              & $+0.0$ & $+5.6$ & $+0.6$ & $+18.6$ & $+0.4$ \\
Pharmaceutical Science            & $-0.0$ & $+6.1$ & $-0.2$ & $+17.7$ & $+1.2$ \\
\bottomrule
\end{tabular}
\caption{Bottom 10 categories ranked by $|\Delta\text{Semantic Similarity}|$ (Mode~\protect\circled{A}$-$\protect\circled{B}; \texttt{GPT-5.4}). Despite near-zero semantic gaps, writing style ($+4.8$ to $+8.4$) and numeric consistency ($+14.3$ to $+23.1$) remain substantially positive.}
\label{tab:category_gap_bottom10}
\end{table}

\newpage
\section{\sys{} dataset statistics}\label{app:MedConclusion-dataset-statistics}

\paragraph{SJR.} The SJR score quantifies a journal's scientific influence by accounting for both
the volume of citations received and the prestige of the citing sources. It is computed via an iterative algorithm analogous to PageRank: a citation from a highly-ranked journal contributes more to a journal's SJR score than one from a lower-ranked journal, and self-citations are down-weighted to mitigate inflation~\citep{gonzalez2010new}. This design renders SJR size-independent and more robust to citation manipulation than raw impact factors. Because SJR scores are published annually, our dataset captures the longitudinal prestige trajectory of each journal from its earliest available record through 2024.

\subsection{Abstracts' year distribution}

\begin{figure}[h]
\centering
\includegraphics[width=\linewidth]{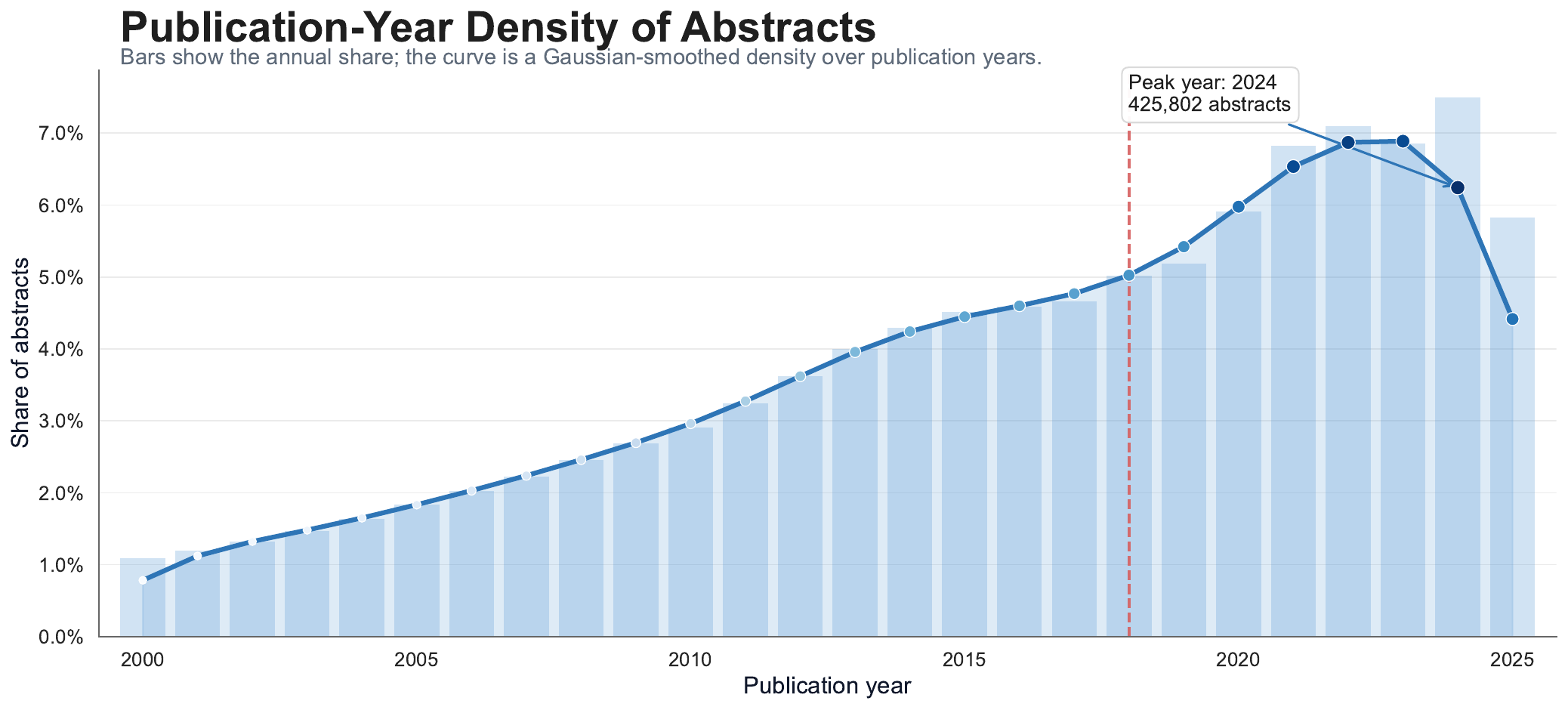}
\caption{Publication-year density of abstracts in \sys{} from 2000--2025. The median publication year is 2018.}
\label{fig:year_distribution}
\end{figure}

\subsection{Journal category \& SJR score distribution}

\begin{figure}[!ht]
\centering
\includegraphics[width=\linewidth]{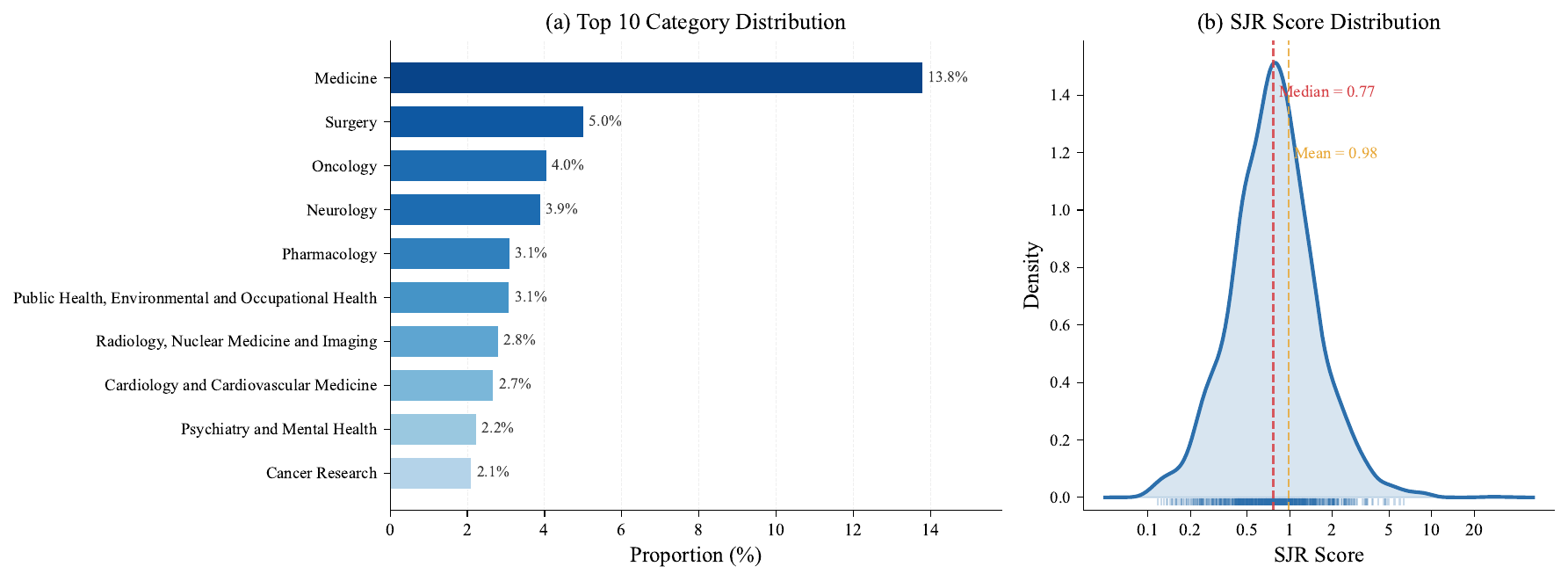}
\caption{Distribution statistics of \sys{}. (a) Top 10 subject categories by proportion of
abstracts in the dataset, with Medicine being the dominant category at 13.8\%. (b) Distribution
of SJR scores across the 3,772 journals (median $= 0.77$, mean $= 0.98$), showing a
right-skewed distribution concentrated in the low-to-moderate prestige range.}
\label{fig:dataset_statistics}
\end{figure}

\newpage
\section{Conclusion Label Variants}
\label{app:conclusion-label-variants}

We use the conclusion label variants shown in Table~\ref{tab:conclusion-label-variants} to identify conclusion-type sections in structured abstracts.

\begin{table}[!htb]
\begin{tcolorbox}[title=Conclusion Label Variants,
  colback=white, colframe=black!60]
\small
\begin{multicols}{2}
\begin{itemize}
    \item \texttt{CONCLUSION}
    \item \texttt{CONCLUSIONS}
    \item \texttt{CONCLUSION(S)}
    \item \texttt{CONCLUSIONS AND RELEVANCE}
    \item \texttt{CONCLUSION AND RELEVANCE}
    \item \texttt{CONCLUSIONS AND IMPLICATIONS}
    \item \texttt{CONCLUSION AND IMPLICATIONS}
    \item \texttt{CONCLUSIONS AND IMPORTANCE}
    \item \texttt{CONCLUSION AND IMPORTANCE}
    \item \texttt{CONCLUSION AND SIGNIFICANCE}
    \item \texttt{CONCLUSIONS AND SIGNIFICANCE}
    \item \texttt{CONCLUSION AND INTERPRETATION}
    \item \texttt{CONCLUSIONS AND INTERPRETATION}
    \item \texttt{CONCLUSIONS AND CLINICAL RELEVANCE}
    \item \texttt{CONCLUSION AND CLINICAL RELEVANCE}
    \item \texttt{CONCLUSIONS AND CLINICAL IMPORTANCE}
    \item \texttt{CONCLUSION AND CLINICAL IMPORTANCE}
    \item \texttt{AUTHORS' CONCLUSIONS}
    \item \texttt{AUTHORS' CONCLUSION}
    \item \texttt{MAIN CONCLUSIONS}
    \item \texttt{MAIN CONCLUSION}
\end{itemize}
\end{multicols}
\end{tcolorbox}
\caption{Conclusion label variants used to identify conclusion-type sections in structured abstracts.}
\label{tab:conclusion-label-variants}
\end{table}

\newpage
\section{Model configurations}\label{sec:models}


\begin{table}[!hbt]
\centering
\scriptsize
\setlength{\tabcolsep}{3pt}
\resizebox{\linewidth}{!}{%
\begin{tabular}{llccccc}
\toprule
Short Name & Full Name & Access & Capability & Scale & \shortstack{Max New\\Tokens} & Temp. \\
\midrule
\multicolumn{7}{c}{\textbf{General-purpose Models}} \\
\midrule
GPT-5.4 & gpt-5.4 \citep{openai_gpt54_2026} & Proprietary & General-purpose & Large & 1024 & 0 \\
Gemini 3.1 Pro & gemini-3.1-pro \citep{gemini31pro_modelcard_2026}  & Proprietary & General-purpose & Large & 1024 & 0 \\
Gemini 3 Flash & gemini-3-flash \citep{gemini3flash_modelcard_2025} & Proprietary & General-purpose & Medium & 1024 & 0 \\
DeepSeek-V3.2 & DeepSeek-V3.2 \citep{deepseekai2025deepseekv32} & Proprietary & General-purpose & Large & 1024 & 0 \\
MiniMax-M2.1 & MiniMax-M2.1 \citep{minimax_m21_2025} & Proprietary & General-purpose & Large & 1024 & 0 \\
Gemma-3-27B & gemma-3-27b-it \citep{gemma_2025} & Open-weight & General-purpose & Medium & 1024 & 0 \\
Llama-3.1-8B & Llama-3.1-8B-Instruct \citep{grattafiori2024llama3herdmodels} & Open-weight & General-purpose & Small & 1024 & 0 \\
Gemma-2-9B & gemma-2-9b-it \citep{gemma_2024} & Open-weight & General-purpose & Small & 1024 & 0 \\
Qwen2.5-7B & Qwen2.5-7B-Instruct \citep{qwen2.5} & Open-weight & General-purpose & Small & 1024 & 0 \\
Qwen3-4B & Qwen3-4B-Instruct-2507 \citep{qwen3technicalreport} & Open-weight & General-purpose & Small & 1024 & 0 \\
Llama-3.2-1B & Llama-3.2-1B-Instruct \citep{meta_llama32_2024} & Open-weight & General-purpose & Small & 1024 & 0 \\
\midrule
\multicolumn{7}{c}{\textbf{Reasoning Models}} \\
\midrule
Kimi-K2 & Kimi-K2-Thinking \citep{kimiteam2026kimik2openagentic} & Open-weight & Reasoning & Large & 1024 & 0 \\
DeepSeek-R1 & DeepSeek-R1 \citep{deepseekai2025deepseekr1incentivizingreasoningcapability} & Open-weight & Reasoning & Large & 1024 & 0 \\
\midrule
\multicolumn{7}{c}{\textbf{Vision-Language Models}} \\
\midrule
GLM-4.6V & GLM-4.6V \citep{vteam2025glm45vglm41vthinkingversatilemultimodal} & Proprietary & Vision-language & Large & 1024 & 0 \\
Qwen2.5-VL-7B & Qwen2.5-VL-7B-Instruct \citep{qwen2.5-VL} & Open-weight & Vision-language & Small & 1024 & 0 \\
\bottomrule
\end{tabular}%
}
\caption{Model metadata and run-configuration summary for all evaluated models. For general-purpose models that expose a \texttt{thinking} mode, \texttt{thinking=none} was used.}
\label{tab:model_metadata_run_config}
\end{table}

\newpage
\section{The use of Large Language Models (LLMs)}
\label{app:llm-usage}
LLM is used only to aid writing quality (proofreading and polishing grammar). No ideas, claims, methods, results, or references are generated by LLMs. All content decisions and revisions are made by the authors.

\end{document}